\documentclass{tADR2e}

\RequirePackage[2019-10-01]{latexrelease}

\makeatletter
\providecommand\@bibsetup[1]{} 
\makeatother

\begin{document}

\jvol{00} \jnum{00} \jyear{2020} \jmonth{June}

\articletype{FULL PAPER}

{
\centering
\Large \bfseries Evaluation of Impression Difference of a Domestic Mobile Manipulator with Autonomous and/or Remote Control in Fetch-and-Carry Tasks\par
\vspace{1cm}
\large Takashi Yamamoto$^{a,c}$$^{\ast}$, Hiroaki Yaguchi$^{b}$, Shohei Kato$^{c}$, and Hiroyuki Okada$^{d}$\par
\vspace{0.5cm}
\small
$^{a}${\em{Frontier Research Center, Toyota Motor Corporation, Toyota, Aichi, Japan}};\par
$^{b}${\em{Kushinada Tech. Co., Ltd., Toshima, Tokyo, Japan}};\par
$^{c}${\em{Graduate School of Engineering, Nagoya Institute of Technology, Nagoya, Aichi, Japan}};\par
$^{d}${\em{Faculty of Engineering, Tamagawa University, Machida, Tokyo, Japan}}\par
\vspace{0.5cm}
$^\ast$Corresponding author. E-mail: tyamamoto@mail.toyota.co.jp\par
\vspace{1cm}
}

\begin{abstract}
Various studies on autonomous robots that perform physical tasks in living environments are being conducted at present to solve the social problems of an aging society with a low birth rate. However, because it remains difficult for all robots to work autonomously in complicated environments, remote control robots may be useful for practical applications. As the purpose of remote control robots is to make direct connections between the operator and user, they may reduce affinity between the robot  and  the user. In this study, we define autonomous remote control robots with the aspects of both autonomy and remote operation explicitly. Affinity is compared among autonomous remote control robots, remote control robots, and autonomous robots, through controlled experiments on interactions in fetch-and-carry tasks using a domestic mobile manipulator, Human Support Robot, on a test field that is compliant with the World Robot Summit 2020 Rulebook. The results demonstrate that affinity exhibits the following order: autonomous robots, autonomous remote control robots, and remote control robots. Considering that autonomous robots remain under development, autonomous remote control robots may be beneficial for expanding the abilities of physical tasks and social acceptance.
\medskip

\end{abstract}

\section{Introduction}
\label{sec:intro}
\begin{figure}
\begin{center}
\includegraphics[width=100mm]{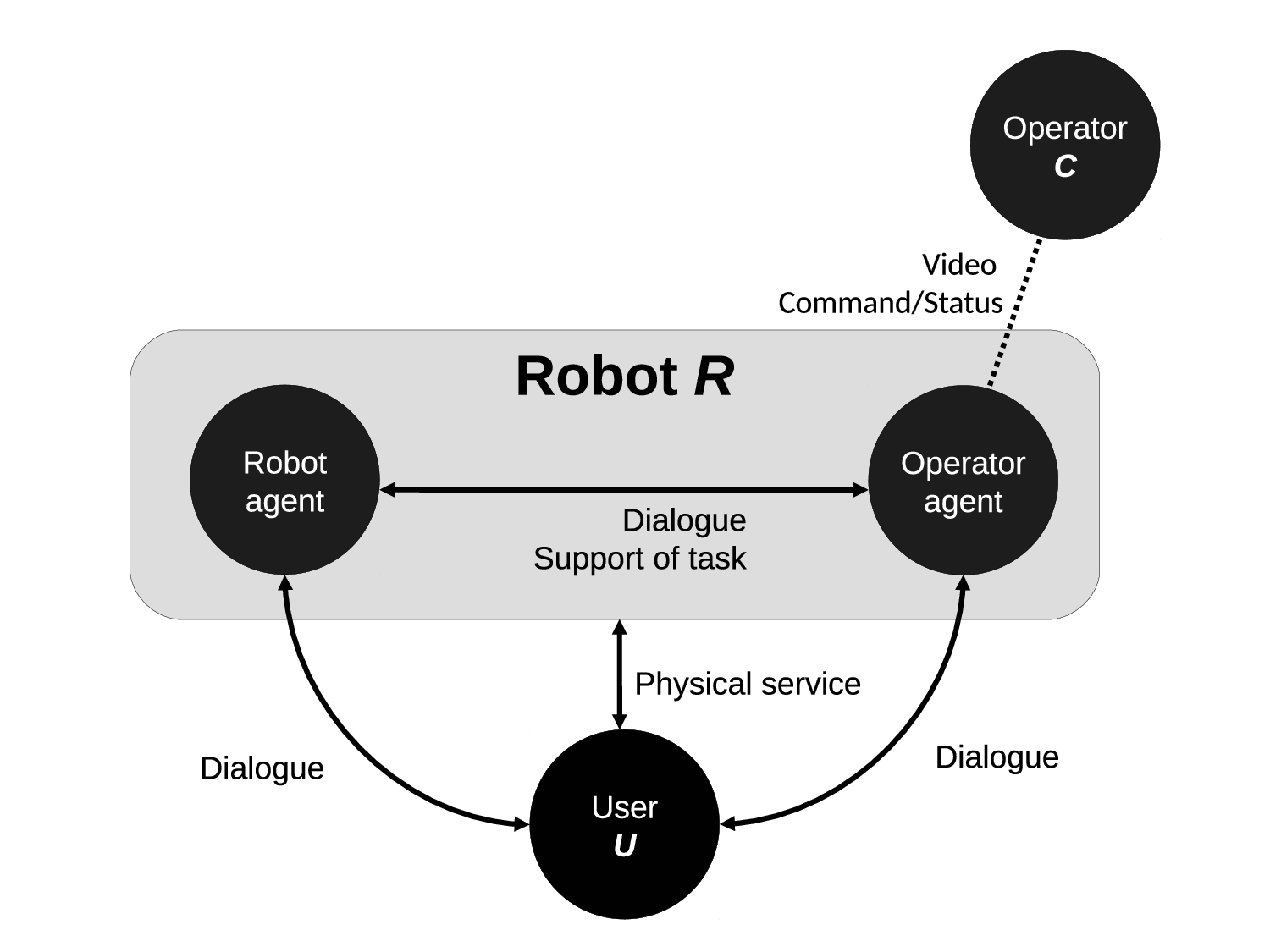}
\caption{Configuration of autonomous remote control robot}
\label{fig:3Persons}
\end{center}
\end{figure}

The development of robots that autonomously perform physical tasks in environments with people (autonomous robots) has been actively researched \cite{bohren2011towards,borst2009rollin,Robomech2019,watanabe2013cooking}. Autonomous robots may provide a solution for various social problems, such as labor force shortages owing to the aging population with a low birth rate, release from monotonous or harsh work for people, support independent living for elderly or disabled people, and housework support for busy people. However, because it remains difficult for robots to perform all tasks autonomously in complex environments, robots that are remotely controlled by operators (remote control robots) are considered to be useful for practical applications. Moreover, as the data from remote operations are available for the learning of robots, remote control robots may provide an effective means to promote autonomous robot technology \cite{iwata2018learning}. However, for such robots to be widely used in people's lives, it is important to improve the social acceptance of robots \cite{duffy2003anthropomorphism,heerink2008influence,gaudiello2016trust,klamer2010acceptance}. The relationship between robots and humans has been regarded as one of the key points for social acceptance. Within the above context, we study the effects of autonomous and remote control functions on the relationship between user and robot. In this paper, we refer to the affinity for the robot, which means that the robot makes the user feel friendliness, a positive impression, and a good relationship. We assume that a higher affinity with users will result in the improvement of social acceptance to a greater extent. As noted previously, at present, remote control robots are considered to be more practical and useful than autonomous robots in many physical tasks that cannot be completed by robots autonomously. However, as the relationship between remote operator and user is centered on the remote control robot, the presence of the robot may be obscured for the user. As a result, the affinity between the user and the robot may be reduced. In this study, we present a new robot configuration (the autonomous remote control robot; Figure~\ref{fig:3Persons}) in addition to the remote control and autonomous robots. We perform a comparative study of the affinity for robots among these three configurations with physical entities.

The autonomous remote control robot is defined as a robot that has both the presence of the remote operator (operator agent) and the autonomous robot (robot agent) inside of it, as shown in Figure~\ref{fig:3Persons}.
Both the operator agent and robot agent exist explicitly and share knowledge and instructions regarding physical tasks through interactions among three parties, namely the above two agents and the user who receives the services.
To explain Figure~\ref{fig:3Persons}, for example, we assume a case where user $U$ asks robot $R$ to bring them something to drink. When $U$ asks $R$ to do the task, the robot agent confirms and replies as a dialogue with $U$. If the task is difficult to perform autonomously, the robot agent asks the operator agent for assistance as an explicit dialogue for $U$. In case the robot agent does not know where to go (e.g., kitchen), the robot agent asks the operator agent, and the operator agent tells the robot agent about the name of the place (kitchen) as an explicit dialogue for $U$. Or, if the robot agent cannot grasp the target object, the robot agent asks the operator agent for support as an explicit dialogue for $U$ and the operator agent takes control of the physical body of $R$ and operates it explicitly for $U$. The robot agent and the operator agent share one common physical body and control it  based on arbitration through explicit dialogue for U, while they and the user communicate mutually to bring something to drink as a physical service to $U$.

Numerous examples of remote control robots with autonomous functions to support the operations of remote operators are available. However, as mentioned previously, the concept and purpose of the autonomous remote control robot differ from these. It can be stated that new value is offered by the configuration of the autonomous remote control robot if higher affinity can be obtained than previous remote control robots. Furthermore, if autonomous robots can achieve the same or higher affinity than autonomous remote control robots, it is reasonable to anticipate that autonomous remote control robots will first be implemented in society, and their autonomous functions will be improved sequentially to maintain and develop both abilities for physical tasks and the affinity for users simultaneously. To demonstrate this, we implemented remote control functions, autonomous functions, and autonomous remote control functions in a domestic mobile manipulator. Under the same environment, with the same hardware and tasks with the same purpose, we confirmed the difference between their affinities and which configuration exhibited the highest affinity through subject experiments. Robots that combine autonomous functions and remote control functions to perform physical tasks have been developed continuously. For example, a system was developed that performs fetch-and-carry tasks (FCTs) by tapping or handwriting a line on a target object on the screen of a tablet PC \cite{hashimoto2013field}. Furthermore, in subsequent studies \cite{nagahama2018learning}, a robot learning method was realized by teaching simple furniture operations through the tablet. Although autonomous functions were mentioned in these studies, their purpose was mainly to provide easier methods for remote operators to control robots, and the affinity between user and robot was not particularly considered. The research in \cite{ando2018experimental} focused on onomatopoeia emitted by a robot to assist the operation of a remote operator, while that in \cite{tachi1985tele} proposed comprehensive sensory sharing through a remote robot. In both cases, the relationship between the remote operator and the robot was the main focus. Various studies have been conducted on communication between robots and humans. For example, research on single robot versus single user \cite{kato2004development}, single robot versus multiple users \cite{Matsusaka2001,Fujie2012}, and multiple robots versus single user \cite{Kanda2002} has been presented. However, these works mainly focused on autonomous communication robots and not robots that perform physical tasks. Moreover, the explicit coexistence of the operator and autonomy of the robot was not studied. In \cite{kanda2001e}, it was experimentally verified that a user feels the presence of a remote operator in dialogues between the remote operator and user via the robot, even if autonomous dialogues are inserted in the middle. However, the user impression of physical robots with the agent and remote operator was not evaluated. In \cite {yamaoka2007e}, it was explained that, when a subject interacts with a remote-controlled robot through dialogue or body contact, the subject tends to sense interaction with the robot itself, and not the remote operator, even if the information is received in advance. However, the main function of this robot was verbal and physical contact communication and not physical task executions. Furthermore, the explicit coexistence of the robot and operator agents was not studied. The study of \cite {ogawa2008itaco} demonstrated that an agent could move to various devices, including robots, so that the user can interact smoothly with the devices. However, the user impression of physical robots with the agent and remote operator was not evaluated.

In this paper, we aim to clarify the impression difference among remote control, autonomous, and autonomous remote control robots with FCTs in the home environment. We assume that the affinity of autonomous remote control robots is higher than that of remote control robots and less than or equal to that of autonomous robots. If the above assumption holds, autonomous remote control robots could be practical and beneficial in terms of compatibility between social acceptance and task performance, as mentioned earlier in this section.

The remainder of this paper is organized as follows. Section~\ref{sec:task_system} describes the robot, target task, and implemented system. In this study, we consider an FCT, which is the basis of all physical tasks. We attempt to define the conditions of the task as clearly as possible for future research. In Section~\ref{sec:experiment}, we explain the design of the subject experiments to compare autonomous remote control functions with remote control functions and autonomous functions. We evaluate the results using two methods, such as the semantic differential (SD) method and ranking method. In Section~\ref{sec:result}, we present the experimental results. With the data from the subjects, we objectively determine the factors related to the affinity. In Section~\ref {sec:discussion}, we discuss the results, and in Section~\ref {sec:conclusion}, we provide the conclusions of the study.

\section{System Design and Target Task}
\label{sec:task_system}

\subsection{Robot and Control Mode}
\label{sec:struct_robot}

\begin{figure}
\begin{center}
\includegraphics[width=100mm]{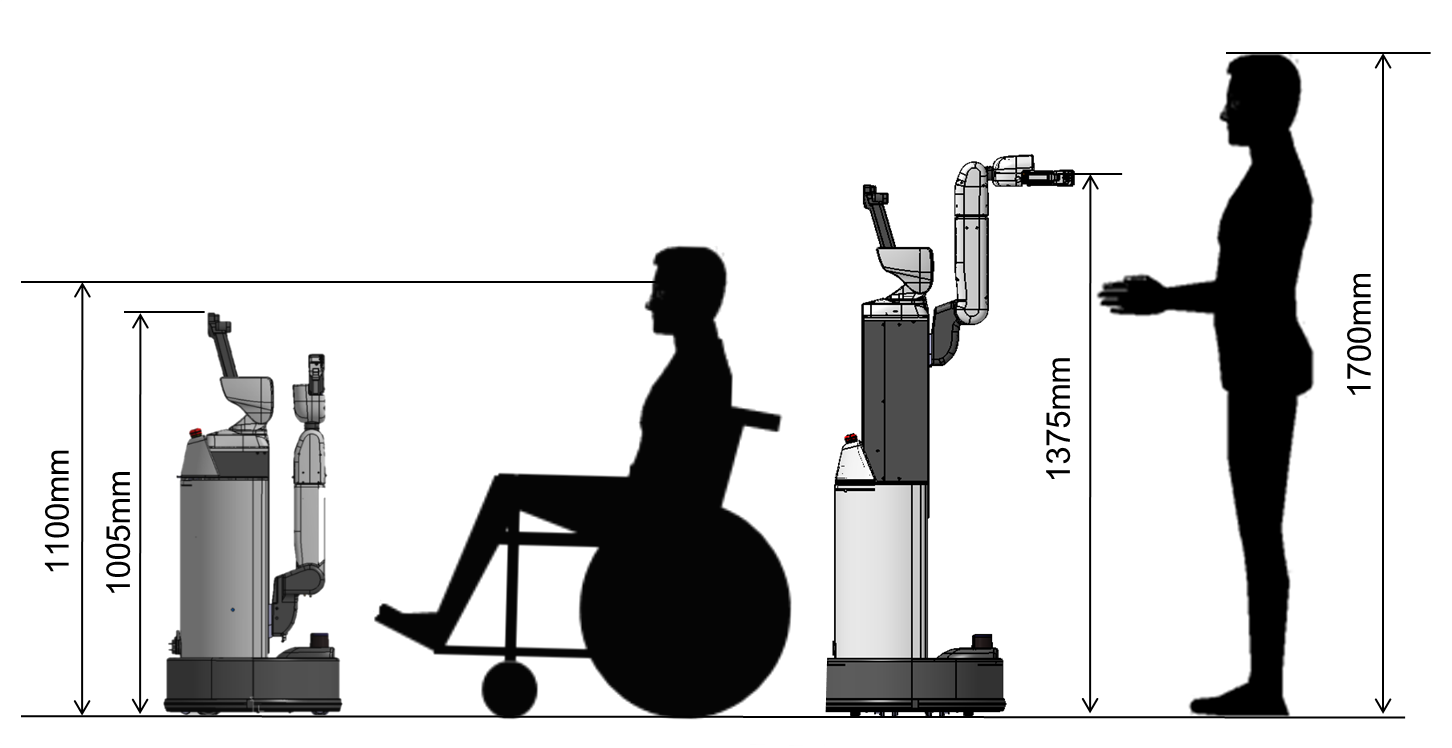}
\caption{HSR}
\label{fig:HSR-Human}
\end{center}
\end{figure}

We used the Human Support Robot (HSR) \cite{Robomech2019} in Figure~\ref{fig:HSR-Human} for this study. The HSR is the research platform of a domestic mobile manipulator with functionality, safety, and a small physique for performing physical tasks in living environments. It has been used in various research projects and international robot competitions \cite{yamamoto2019human,berenstein2018robustly, yi2019mobile, pena2018eeva, quispe2017learning, yamaguchi2019live, itadera2019impedance}.

In this study, we implemented the following three control modes, selectively on the same HSR.
\begin{description}
  \item[M1]: Remote control mode:\\ $R$ performs physical tasks by the remote control of $C$.
  \item[M2]: Autonomous control mode:\\  $R$ performs physical tasks autonomously by commands from $C$ or $U$.
  \item[M3]: Autonomous remote control mode:\\  $R$ performs physical tasks by both the remote control of $C$ and autonomous control, as per the autonomous remote control robot illustrated in Figure~\ref{fig:3Persons}.
\end{description}

We denote $R$ as the robot, $E$ as the environment in which $R$ performs tasks, $U$ as the user who receives services from $R$, $C$ as the operator who controls $R$ remotely, $O$ as the target object, which $R$ fetches and carries to $U$, $P$ as the place in which $O$ exists, and $\Omega_{e}=\{U, C, R\}$ as the set of the three parties. In real cases, $C$ and $U$ may even be the same person, whereby $U$ receives the service of $R$ by own remote control.

The modes M1, M2, and M3 corresponds to remote control, autonomous and autonomous remote control robots, respectively.

\subsection{Target Task}
\label{sec:target_task}

Although the FCT is a simple and basic task, diverse cases exist depending on the conditions. Firstly, we classify the extensive conditions of the FCT and clarify the scope of this research.
\subsubsection{Definition of Task}
\label{sec:def_task}

We focus on an FCT in which $R$ fetches and carries $O$ to $U$ because it includes major technical elements such as movement, recognition, and grasping; moreover, it can form the basis of any complex physical task. Next, we divide the FCT into sub-tasks (STs), as follows:

\begin{description}
  \item[ST1]: Go to $P$
  \item[ST2]: Take $O$
  \item[ST3]: Go to $U$
  \item[ST4]: Hand $O$ to $U$
\end{description}

\begin{figure}
\begin{center}
\includegraphics[width=100mm]{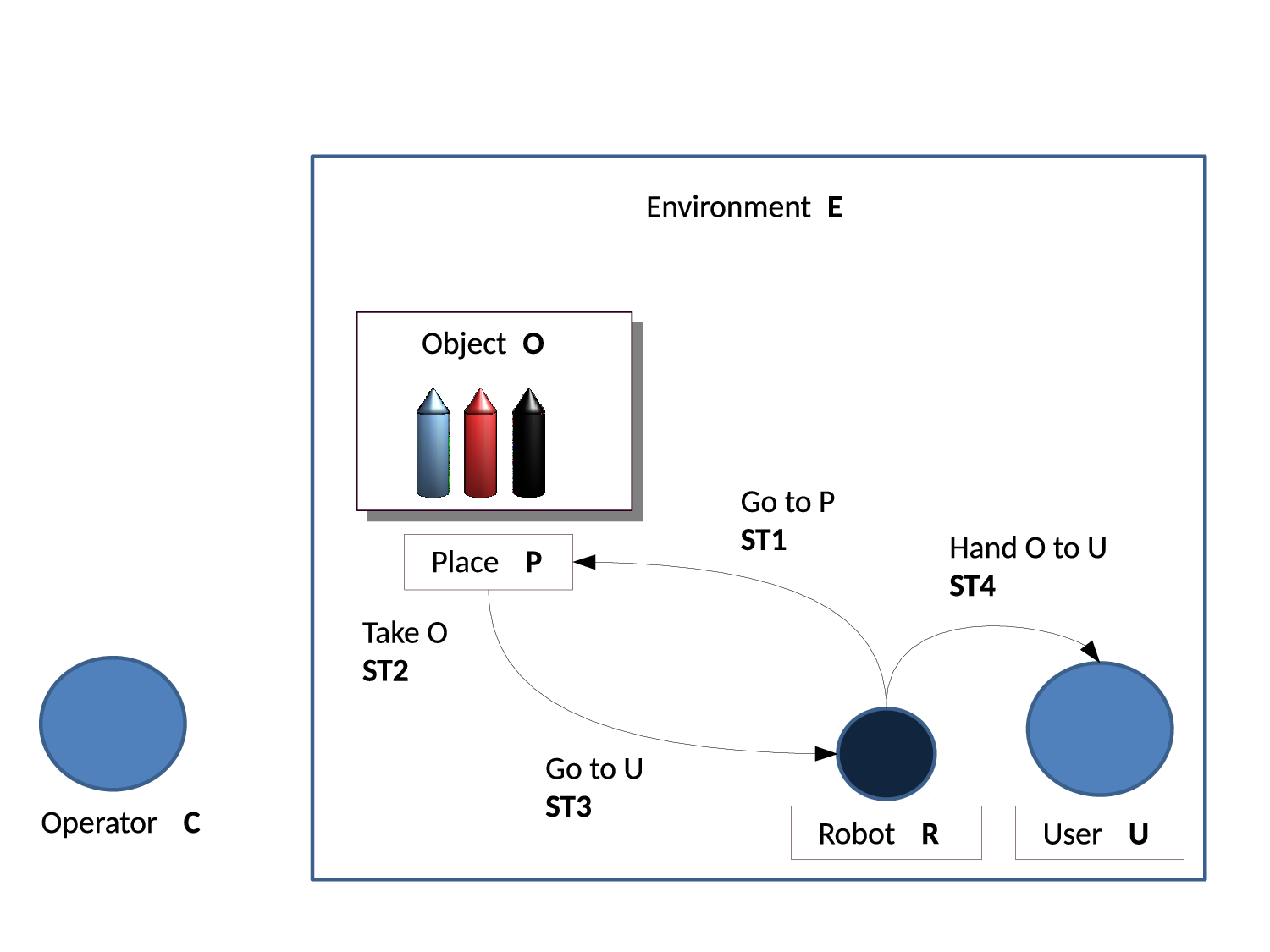}
\caption{STs of FCT}
\label{fig:Subtask}
\end{center}
\end{figure}

As illustrated in Figure~\ref{fig:Subtask}, each ST is sequentially executed in the FCT. Specifically, $R$ moves to $P$ (ST1), takes $O$ (ST2), moves to $U$ (ST3), and hands $O$ to $U$ (ST4).

We first clarify the combination of knowledge and abilities of $U$, $C$, and $R$ generically. For example, it is difficult for $R$ to move autonomously in ST1 and ST3 if $R$ does not know the information of $E$. In fact, numerous combinations of the knowledge and abilities of $U$, $C$, and $R$ exist. We define $\Omega_p$ as the set of all combinations of knowledge and abilities thereof. One element $i \in \Omega_p$ is denoted by `pattern $i$'. Furthermore, we define $\Omega_k=\{K_O,K_P,K_M,K_G\}$ as the set of four types of knowledge and abilities and a two-valued logic function (referred to as the `condition function') on pattern $i$, $L_i(x,y)$ ($x\in \Omega_e=\{U,C,R\}$, $y\in \Omega_k$), as follows:
 
 \begin{itemize}
  \item $y=K_O$: Initial knowledge of object  \\
  In the case where $x$ knows $O$ initially or decides $O$ by itself, the condition function $L_i(x,K_O)=1$ (TRUE). If not, $L_i(x,K_O)=0$ (FALSE). 
  
 In other words, $K_O$ represents the initiative to decide the target object $O$ of the task. Even if $L_i(x,K_O)=0$, it is possible for $x$ to know $O$, after hearing what $O$ is from the others assuming $x$ can understand what $O$ is. We assume that $L_i(R,K_O)=1$ means that $R$ knows what to grasp from the beginning of the task without hearing from $U$ or $C$  and it does not mean that $R$ has the ability to detect $O$ in $E$. The ability is included in the condition of $K_G$ mentioned below.

  \item $y=K_P$: Initial knowledge of place\\
  In the case where $x$ knows where $P$ is initially, the condition function $L_i(x,K_P)=1$ (TRUE). If not, $L_i(x,K_P)=0$ (FALSE). Even if $L_i(x,K_P)=0$, it is possible for $x$ to share the information of $P$, after hearing $P$ in the case where $x$ has the map of $E$ with $P$.
  \item $y=K_M$: Ability of movement\\
  In the case where $x$ can cause $R$ to move to any acceptable goal in $E$ by its order or control, the condition function $L_i(x,K_M)=1$ (TRUE). If not, $L_i(x,K_M)=0$ (FALSE).   
It is possible for $R$ to move autonomously if $L_i(R,K_M)=1$.
  \item $y=K_G$: Ability of grasping\\
    In the case where $x$ can cause $R$ to grasp any acceptable object by its order or control, the condition function $L_i(x,K_G)=1$ (TRUE). If not, $L_i(x,K_G)=0$ (FALSE).   
We assume that it is possible for $R$ to grasp any acceptable object autonomously along with the ability to detect it in $E$ in case that $L_i(R,K_G)=1$.
\end{itemize}

In a general FCT, there are 4096 patterns ($ = (2 ^ 4) ^ 3 $) based on simple combinations of the four conditions of $\Omega_k=\{K_O,K_P,K_M,K_G\}$. However, the number of patterns that should be evaluated can be decreased by setting the following reasonable assumptions:

\begin{description}
  \item[a)] Only one of $U, C, R$ has the knowledge and abilities of $\Omega_k=\{K_O,K_P,K_M,K_G\}$.\\
   In this case, the following condition is satisfied:
  \begin{equation}
     \label{Condition1}
       \forall  y\in \Omega_k,  \ L_i(U,y)  \oplus  L_i(C,y) \oplus  L_i(R,y) =1.
  \end{equation}  
 Owing to this condition, the redundancy of the roles of $U$, $C$, and $R$ is eliminated.
  \item[b)] $C$ or $R$ has the ability of movement if it has knowledge of the place.\\
  In this case, the following condition is satisfied:
  \begin{equation}
     \label{Condition2}
     \forall x\in\{C,R\},  \ L_i(x,K_P) \to L_i(x,K_M) =1.
  \end{equation}  
 
We can assume that $C$ or $R$ has the map of $E$ if it has knowledge of the place. This means that $C$ can control $R$ or $R$ can move autonomously using the map of $E$.
\end{description}

\subsubsection{Classification of Pattern for Control Mode}
\label{sec:control_mode}

We can classify each pattern $i$ as a control mode, depending on the ability of $R$, as follows:

\begin{itemize}
\item Classification condition of remote control mode (M1): \\
   If the following condition is satisfied, the pattern $i$ is classified as M1 because $R$ cannot move and grasp objects autonomously.
  \begin{equation}
   \label{M1}
   \lnot(L_i(R,K_M) \lor L_i(R,K_G))=1.
  \end{equation}
  
\item Classification condition of autonomous mode (M2):\\
   If the following condition is satisfied, the pattern $i$ is classified as M2 because $R$ can move and grasp objects autonomously.
  \begin{equation}
      \label{M2}
   L_i(R,K_M) \land L_i(R,K_G)=1.
  \end{equation}
  
  \item Classification condition of autonomous remote control mode (M3):\\
         If the following condition is satisfied, the pattern $i$ is classified as M3 because $R$ can only perform either autonomous movement or autonomous grasping.
  \begin{equation}
         \label{M3}
    L_i(R,K_M) \oplus L_i(R,K_G)=1.
  \end{equation}
  
\end{itemize}

\subsubsection{Selection of User and Evaluation Task}
There are various cases for the conditions of user $U$. In the evaluation of this research, we select the following conditions: 
     \begin{equation}
     \label{U1}
      L_i(U,K_O)=1,\\
     \end{equation}   
      \begin{equation}
           \label{U2}
          \forall y\in\{K_P,K_M,K_G\}, \  L_i(U,y)  =0.
      \end{equation}
      
The above conditions mean that $U$ only orders $O$, whereas $C$ controls $R$ remotely or $R$ acts autonomously. We denote the subset $\Omega_p^s$ of patterns satisfying the conditions of equations (\ref{Condition1}), (\ref{Condition2}), (\ref{U1}), and (\ref{U2}) as $\Omega_p^s \subset  \Omega_p$. As a result, four patterns exist in $\Omega_p^s$, which is denoted as $\Omega_p^s=\{PN_1,PN_2,PN_3,PN_4\} $. Table~\ref{tbl:condition} displays the values of the condition functions and the control mode classified by (\ref{M1}), (\ref{M2}), and (\ref{M3}) for each pattern in $\Omega_p^s$.

\begin{table}
\tbl{Value of condition function and control mode for each pattern}
{
\begin{tabular}{l|cccc|cccc|cccc|c}
\hline
\hline
 & \multicolumn{4}{c}{U} &  \multicolumn{4}{c}{C}&  \multicolumn{4}{c}{R} &  \\
 \hline
  & $K_O$ & $K_P$ &$K_M$ &$K_G$ & $K_O$ & $K_P$ &$K_M$ &$K_G$ & $K_O$ & $K_P$ &$K_M$ &$K_G$ & Mode \\
 \hline
PN1  & 1 & 0 & 0 & 0 & 0 & 1 & 1 & 1 & 0 & 0 & 0 & 0 &  M1\\
PN2  & 1 & 0 & 0 & 0 & 0 & 1 & 1 & 0 & 0 & 0 & 0 & 1 &  M3\\
PN3  & 1 & 0 & 0 & 0 & 0 & 0 & 0 & 1 & 0 & 1 & 1 & 0 &  M3\\
PN4  & 1 & 0 & 0 & 0 & 0 & 0 & 0 & 0 & 0 & 1 & 1 & 1 &  M2\\
\hline
\hline
\end{tabular}
}
\label{tbl:condition}
\end{table}

We describe specific examples that match the conditions of $\Omega_p^s$ to ensure that these patterns are realistic, as follows:

\begin{description}
  \item[$PN_1$]: In this case, $C$ has $K_P$, $K_M$, and $K_G$ and the control mode is M1. As an example, there is a case in which helper $C$ fully supports $U$ with limb disorders, with remote control of $R$ at $U$'s home.
  \item[$PN_2$]: In this case, $C$ has $K_P$ and $K_M$, and $R$ has $K_G$. The control mode is M3. As an example, we assume the following situation. Elderly $U$ starts to reside at a care home for the first time with his robot $R$ and asks $R$ to bring a product $O$, which $R$ knows. Here, $R$ does not know $E$ and $P$. The remote helper $C$ can only make $R$ go to the location, following which $R$ can find $O$ and take it.
  \item[$PN_3$]: In this case, $R$ has $K_P$ and $K_M$, and $C$ has $K_G$. The control mode is M3. As an example, we assume the following situation. Helper $C$ remotely supports elderly $U$ at $U$'s house for the first time. $U$ requests $O$ from $R$, and $R$ moves to $P$ autonomously. However, $R$ cannot detect and grasp $O$ because numerous similar objects and obstacles exist in $P$. Because $R$ cannot perform autonomously, $C$ remotely controls $R$ to support it.
  \item[$PN_4$]: In this case, $R$ has $K_P$, $K_M$, and $K_G$ and $R$ is a typical autonomous servant robot that can fetch and carry any accepted $O$ requested by any $U$, including a healthy person, autonomously. The control mode is M2.
\end{description}
  
\subsection{System Design}

\subsubsection{Selection of Remote Control Method}

There are various means of achieving the remote control of robots, particularly for grasping objects, as explained in the following examples.
\begin{itemize}
\item Real-time instruction \cite{tachi1985tele,ando2018experimental}\\
In this method, $C$ sends instructions of each movement to $R$ in real-time. In general, remote control devices such as joysticks, wearable devices that measure the movement of $C$, and motion measurement sensors installed in the environment are required. If a relationship between humans replaces this method, it would be similar to the method of direct instruction by taking the opponent's hand and transmitting the action.
\item Intermittent trajectory instruction \cite{2011touchmeE,2009sketch}\\
In this method, $C$ sends instructions of the target trajectories to $R$ intermittently. Once $C$ provides initial rough information of the trajectories of $R$, $R$ may modify them, adapting to the environment autonomously. For example, $C$ indicates the target trajectory by finger operation or handwriting on the tablet screen, and $R$ determines the actual trajectory on this basis. If this method is replaced by a relationship between humans, it would be similar to the method of gesture instruction by showing the real movement.
\item Goal instruction \cite{hashimoto2013field,nagahama2018learning}\\
When $C$ performs a screen tap or provides a simple drawing instruction on the tablet screen to the target, the trajectories of $R$ are generated, and $R$ starts to move automatically. If this method is replaced by a relationship between humans, it would be similar to the method of saying `Take it' and pointing at the target object with a finger.
\end{itemize}

In this research, we adopt the goal instruction method, which is the easiest to operate by a remote operator and has already been developed on the HSR by the authors.

\subsubsection{User Interface}
\begin{figure}
\begin{center}
\includegraphics[width=100mm]{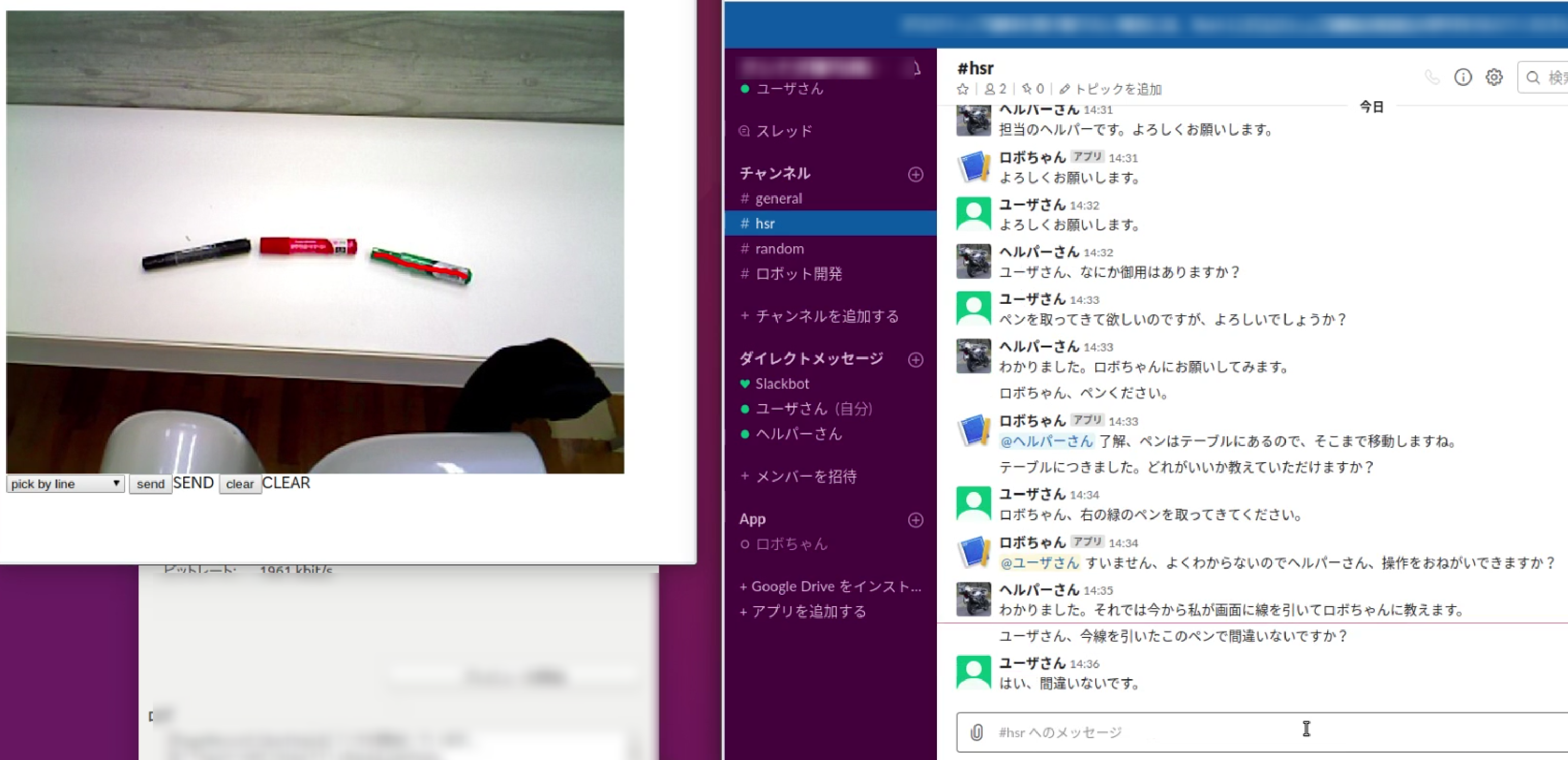}
\caption{User interface}
\label{fig:UserPC}
\end{center}
\end{figure}

In this section, we explain the user interface (UI). In this study, communication among three parties ($U$, $C$, and $R$) from different locations is required. Therefore, instead of speech dialogue, in which recognition failures and conversation collisions are likely to occur, we adopt a text chat UI, which is widely used in devices such as mobile phones. The text chat is used extensively for dialogue, and we believe that it poses no real problem as an alternative to oral communication. For visual instructions and confirmations, both $U$ and $C$ use the image of the robot camera to share visual information. Both $U$ and $C$ use the same UI on each PC. Figure ~\ref{fig:UserPC} presents the UI image displayed on the operation PC. The left side of the screen is the camera image of $R$, while the right side is the text chat.

\subsubsection{Overview of Implemented System}

\begin{figure}[tb]
\begin{center}
\includegraphics[width=100mm]{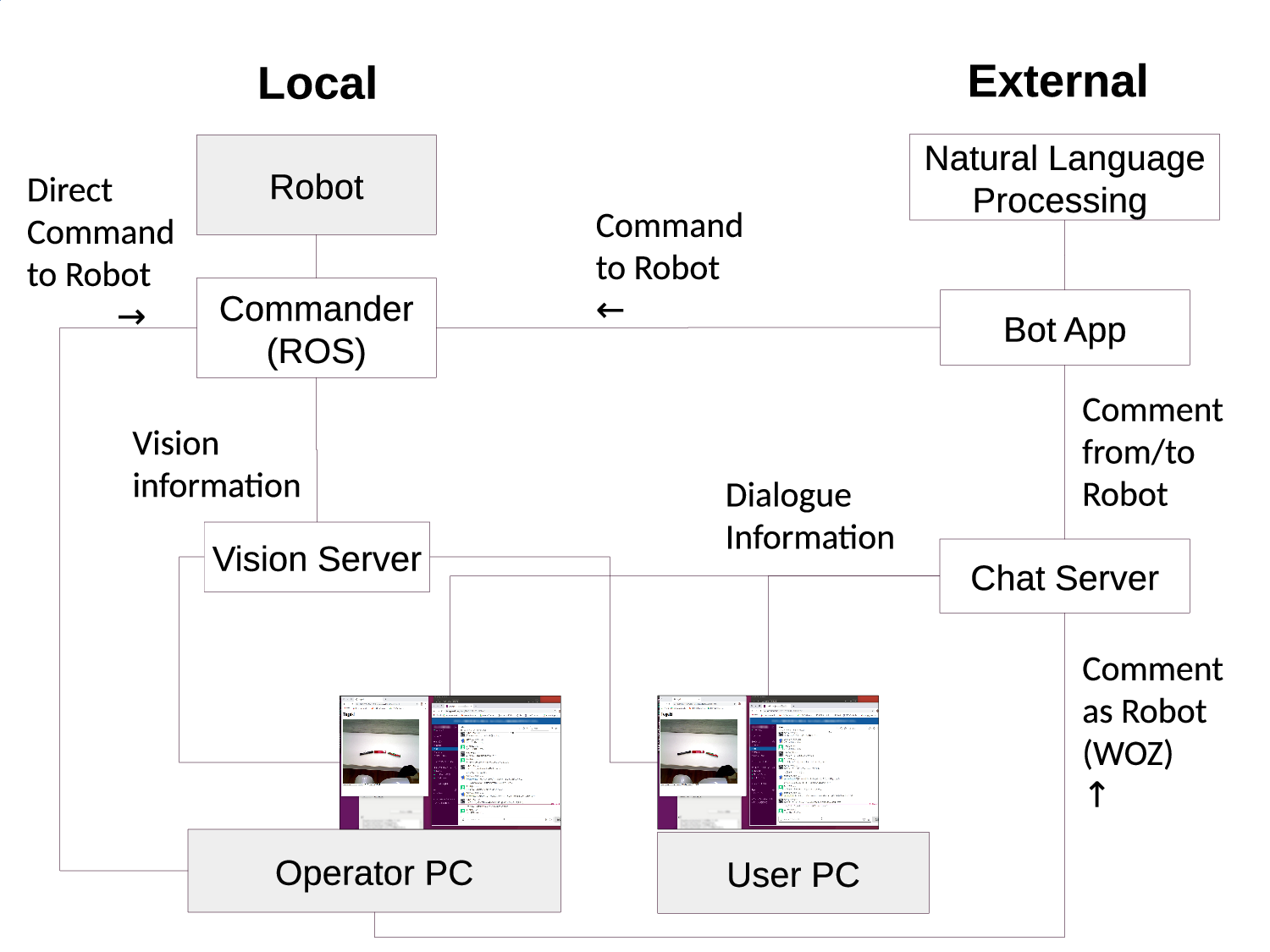}
\caption{System configuration}
\label{fig:System}
\end{center}
\end{figure}

The implemented system configuration is presented in Figure~\ref{fig:System}. Comments of $U$ and $C$ are entered through the Chat Channel of the browser on each PC. These are interpreted by natural language processing (NLP) via the Chat Server and Bot App. In NLP, it is possible to interpret various sentence expressions by learning multiple sentences for each command in advance. The Bot App performs dialogue control based on the interpreted command information. In this case, the states and transitions of the dialogue are represented by a finite state machine. From the Bot App, comments are sent to the Chat Server, commands are sent as an ROS Topic to the robot (HSR) via the Commander, and the operation is executed. The chat system used in this study is Slack, and the NLP is Language Understanding (LUIS) from Microsoft. The robot camera image and drawing information by $C$ are displayed on the user PC (User PC) and remote operator PC (Operator PC) from the front end of the browser via the Vision Server. The drawing information of $C$ can be hidden on the User PC. Moreover, it is possible to send commands to the robot directly, without knowledge of the user, from the remote operator PC and to send sentences to the Chat Server as a robot. Using these mechanisms, Wizard of OZ (WOZ) tests  \cite{riek2012wizard}  by $C$ become possible.

\subsubsection{Implementation of STs}
\label{sec:inp_subtask}

Next, we explain the implementation of each ST of the FCT described in \ref{sec:def_task}. In this study, we focus on a comparative evaluation of three control modes; thus, we do not implement all control modes completely. Each ST has almost the same implementation in any control mode. Different control modes are realized by partially incorporating the WOZ test, and changing the screen or dialog contents presented to $U$.
\begin{itemize}
  \item Implementation of ST1\\
  $R$ stores a map of the environment $E$, and the location $P$ is registered. When $P$ is specified, $R$ autonomously moves to $P$ using the standard two-dimensional simultaneous localization and mapping algorithm.
  \item Implementation of ST2\\ 
   The object $O$ is always treated as an unknown object for $R$. Through the camera mounted on robot $R$, $C$ draws a line on $O$ displayed on the UI screen of the Operator PC and specifies the object to be grasped by $R$. Figure~\ref{fig:Pen} presents the Operator PC screen at the time of remote operation, where a red line drawn by $C$ from the Operator PC is displayed on the green pen on the right. At this time, the three-dimensional position of the drawing line is measured from the information on the depth sensor mounted on $R$'s head. $R$ performs the actual motion automatically after planning the grasping posture of the hand and motion of the entire body.
\begin{figure}
\begin{center}
\includegraphics[width=100mm]{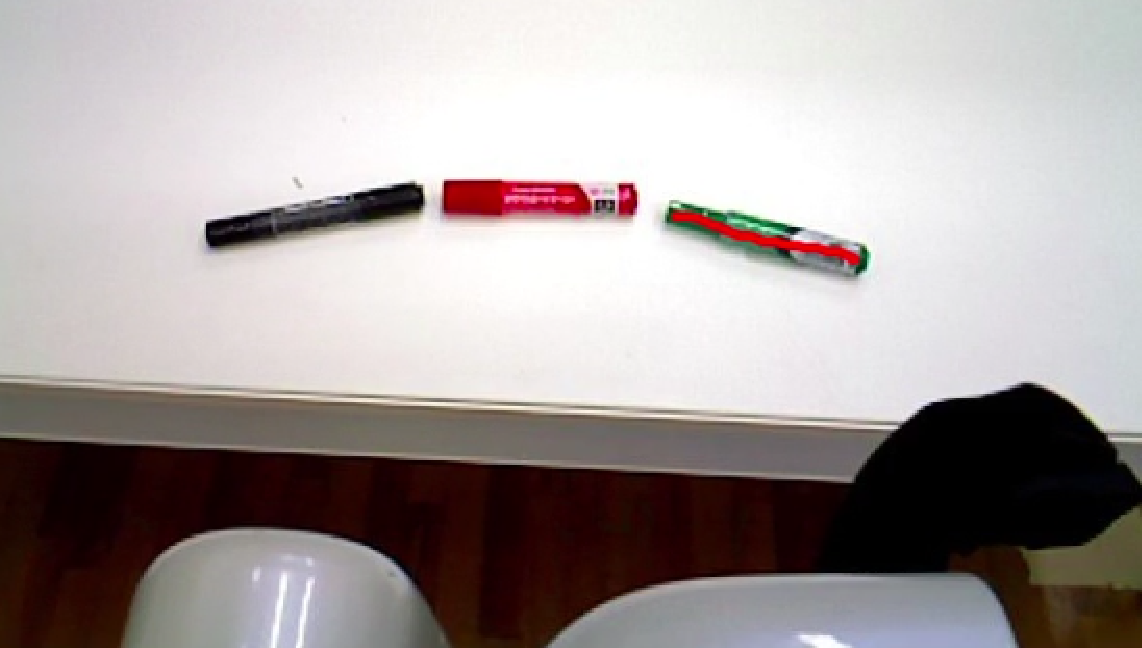}
\caption{Grasping object indication by handwritten line on screen}
\label{fig:Pen}
\end{center}
\end{figure}

 \item Implementation of ST3\\
   Assuming that $U$ is always in the same location, the location of $U$ is registered on $R$’s map in advance. Therefore, the implementation is the same as that of ST1.
  \item Implementation of ST4\\  
  $U$ receives the object $O$ from $R$’s hand after automatically moving $R$'s hand to the fixed position, where it is easy for $U$ to take. $R$ opens the hand after a force sensor mounted on the wrist of $R$ detects the external force generated when $U$ takes $O$ from $R$'s hand.
\end{itemize}

\section{Experiments}
\label{sec:experiment}

\subsection{Overview of Experiments}
\label{sec:abs_exp}

Using the implemented system described in the previous section, interaction experiments on the three control modes by the subjects ($U$), robot $R$, and remote operator ($C$) were performed. Each subject answered questionnaires based on the SD method for each control mode. After the completion of the three experiments with different control modes, the subject was asked to answer the ranking questions for each control mode based on the ranking method. In the experiment, user $U$ was the subject, remote operator $C$ was the developer, and robot $R$ was the HSR. The total number of subjects was 36, consisting of 33 university engineering course students (30 males, from 18 to 22 years old, and 3 females, from 18 to 21 years old) and 3 university officers (3 females, from 41 to 50 years old). The effect of subject demographics will be discussed in subsection \ref{sec:BGandTO}.

\begin{figure}[tb]
\begin{center}
\includegraphics[width=120mm]{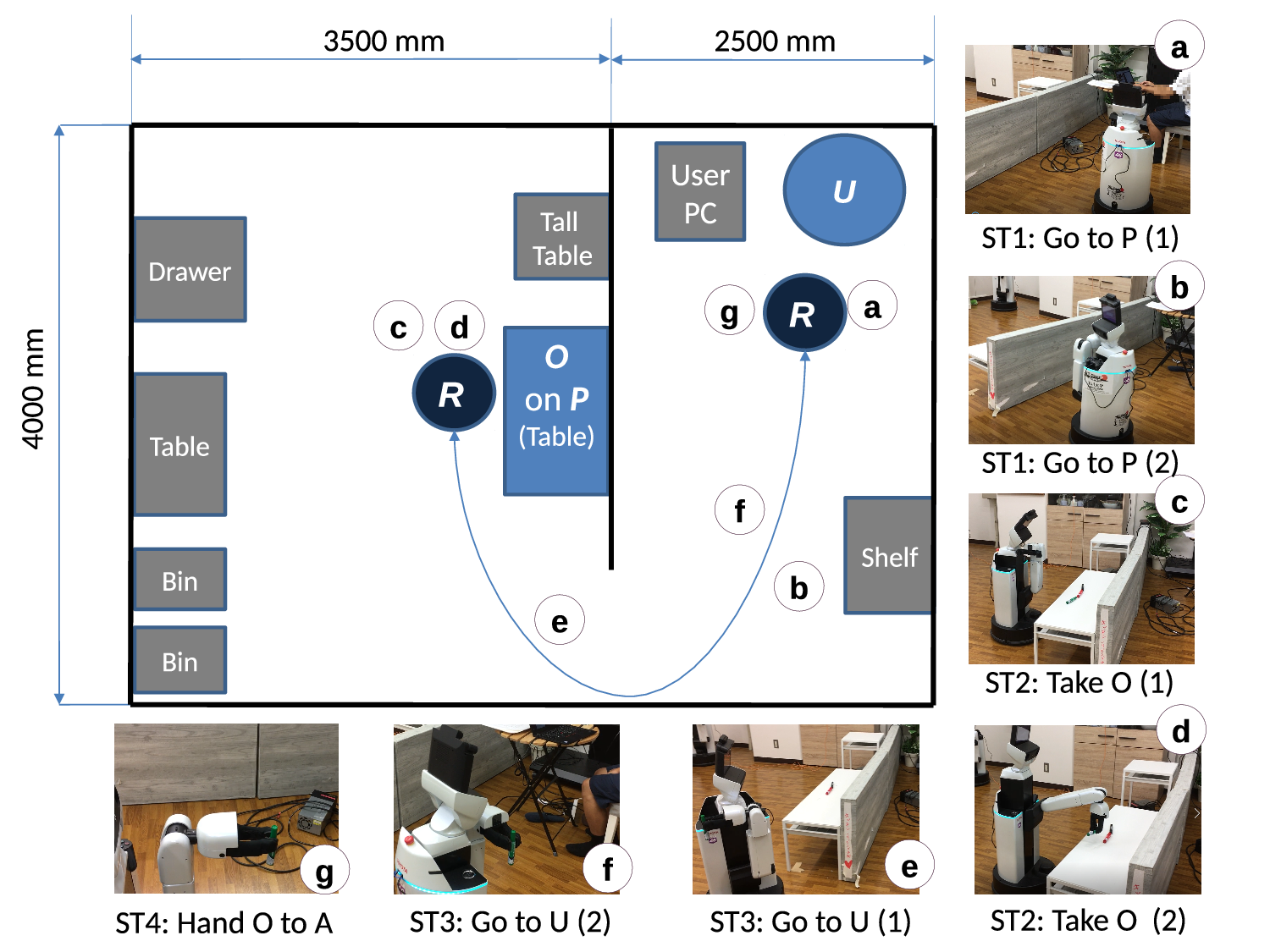}
\caption{Experimental environment based on WRC and execution of STs}
\label{fig:EXP}
\end{center}
\end{figure}

Firstly, the purpose of this experiment was explained to the subjects. At this time, we explained the outline of the three control modes and that each experiment in each mode would be performed sequentially on the same robot in advance. Next, we explained how to use the chat system on the User PC and allowed the subjects to get used to the operation through greetings by chatting with $R$ and $C$. Prior to each experiment, the subject was told which control mode would be used. Even in actual usage scenes, it is natural for the user to know the control mode of the robot in advance. Each experiment was performed by the subject alone, while $C$ remotely controlled $R$ in another room. As each subject performed three experiments of the different control modes (M1, M2, and M3) sequentially, the concern arose that the order of the experiments would affect the results. Therefore, the test was performed by randomly changing the experiment order. Although the number of permutations in the three control modes was 6, each subject was randomly and equally allocated to one of six groups, where each group was in the same order and had six subjects. The groups in the same order were referred to as subject groups A to F (Groups A--F). Influences on the answers by the differences among the subject groups are discussed in subsection \ref{sec:BGandTO}. The experimental environment was based on the World Robot Competition (WRC), Service Robot Category, which will be held at the World Robot Summit (WRS) in October 2020 \cite{okada2019competitions}. As the details of the layout of the competition field are specified in the WRC Rulebook, we expect that additional experiments and comparative studies can be performed under standard environmental conditions. Figure~\ref{fig:EXP} illustrates the outline of the WRC-compliant field and the execution location of the STs described in subsection \ref{sec:inp_subtask}. For the experimental conditions, we assumed that the object $O$ that $U$ wanted to be always a pen. In the following, the contents of the experiment are explained using symbols (a) to (g) of Figure~\ref{fig:EXP}. Three different pens were placed in $P$ in advance. The subject requested $R$ to carry a pen (a), and $R$ moved to the table $P$ where the pens were placed (b). When $R$ arrived at $P$ (c), there were multiple pens on the table, and $U$ selected one pen. When the object was determined, $R$ grasped the selected pen (d) and moved to $U$ (e, f). Upon arriving at $U$, the object $O$ was handed over to $U$ (g). In particular, when selecting a pen, dialogue with $R$ or $C$ was inevitably generated, $U$ finally made its own decision, and $R$ moved according to the intention of $U$. This process is thought to provide a reality of interaction to $U$.

As described above, we explained the prerequisites that the subject wanted the robot to pick up a pen and asked subjects to complete the task honestly together with the others. Apart from this, the subjects were allowed to interact freely with the robot. Although the basic physical behaviors of the robot did not differ depending on the control mode, the actual three-party interactions differed, as indicated below.
\begin{itemize}
  \item Experiment of remote control mode (M1)\\
  This condition is PN1 in Table~\ref{tbl:condition}. An example of the dialogue is presented in Appendix \ref{sec:appendix_dialog1}. $U$ only communicated with $C$ using the chat. In ST1 and ST3, $R$ had to be remotely controlled by $C$, but in fact, $R$ moved autonomously, while $C$ behaved as though controlling $R$. In ST2, $C$ interactively displayed the line drawing of the object instruction on $U$ on the User PC.
  \item Experiment of autonomous control mode (M2)\\
  The condition is PN4 in Table~\ref{tbl:condition}. An example of the dialogue is presented in Appendix \ref{sec:appendix_dialog2}.
   $U$ communicated only with $R$. ST2 was executed as a WOZ test. In particular, $C$ indicated the object with a line drawing on the image and provided operation instructions to $R$, but these processes were not revealed to $U$.
  \item Experiment of autonomous remote control mode (M3)\\
  As indicated in Table~\ref{tbl:condition}, two patterns of $PN_2$ and $PN_3$ were related to M3, but this time only $PN_3$ was implemented. An example of the dialogue is presented in Appendix \ref{sec:appendix_dialog3}. The task was performed with communication among the three parties $U$, $C$, and $R$. The system adequately satisfied the conditions of M3 without any WOZ tests.
\end{itemize}

In this experiment, all actions of $R$ were executed within the field of view of $U$ to make it easier to provide the presence of the robot to $U$ and to allow $U$ to feel the impression of the entire robot, including its hardware. However, interaction with $C$ was restricted to the text chat and robot camera images on the User PC. As mentioned previously, the three control modes had different interaction situations, but the actual robot movements were almost the same. If these conditions produced differences in impressions owing to the differences in the control modes, the main factor was not the difference in the impressions of the robot appearance and movement, or visual expressions of the remote operator. It is assumed that this was owing to the difference in the interaction configuration for the physical task execution. We conclude that these settings are important to extract impression differences as appropriately as possible.
\subsection{Evaluation Methods}
\subsubsection{SD Method}

Every time one experiment was completed, the subject $U$ was asked to answer the adjective pairs presented in Table~\ref{tbl:SD} using a five-step Likert scale. For the SD method, we referred to the adjective pairs in \cite{bartneck2009measurement}, which proposed adjective pairs for the standard evaluation of service robots. As the first step in the analysis, we evaluated one of the most direct adjective pairs (Q6: `Unfriendly--Friendly') for affinity. We determined whether there was a difference in the population mean in each control mode using the $t$-test and Holm's multiple comparison method. If there was no problem with this analysis, factor analysis was performed using the answers to all the questions in Table~\ref{tbl:SD} to gain comprehensive impressions that could not be captured with only one question. When useful factors were extracted, the $t$-test and Holm's multiple comparison method were similarly performed to determine the difference between the population means of the factor scores among the three control modes. In each analysis, we used the effect size of Cohen's $d$, which is used extensively in research, and evaluated the results quantitatively. According to Cohen's criteria, $ d = 0.8, d = 0.5$, and $d = 0.2 $ indicate that the effect is large, medium, and small, respectively \cite{Okubo2012}.
\subsubsection{Ranking Method}

The SD method provided intuitive impressions immediately following the experience of each experiment, while the ranking method provided an impression of looking back on the entire test. Each subject answered the questions in Table~\ref{tbl:Q_Order} after the experiments on the three control modes. In questions 1, 2, and 3, the subjects were requested to answer the ranks from first to third for each control mode. In Q4, the subjects were requested to answer the ranks between the remote control mode and autonomous remote control mode. Q1 concerned the affinity for the robot, which is the central theme of this study. Q2 and Q3 focused on the intellect and sense of security, which are considered as major evaluation factors in addition to the affinity. Q4 was formulated to confirm how the affinity for remote operator $C$ changes depending on the remote control mode (relationship between two parties) or autonomous remote control mode (relationship among three parties). For the analysis of the rank difference owing to the control modes, the Wilcoxon signed-rank test, which is a non-parametric method, and Holm's multiple comparison method were used  and for measuring the effect size $d$, Cliff's $d$ was used \cite{Okubo2012}.

\begin{table}
\tbl{Questions of SD method}{
\begin{tabular}{crl|crl}
\hline
\hline
No. & \multicolumn{2}{c}{Adjective pair} & No. & \multicolumn{2}{c}{Adjective pair} \\
\hline
1 & Unpleasant & Pleasant & 13 & Irresponsible & Responsible \\
2 & Unkind & Kind & 14 & Incompetent & Competent  \\
3 & Inert & Interactive & 15 & Foolish & Sensible  \\ 
4 & Moving rigidly & Moving elegantly & 16 & Fake & Natural \\
5 & Unintelligent & Intelligent & 17 & Dislike  & Like \\ 
6 & Unfriendly & Friendly & 18& Dead & Alive  \\
7 & Stagnant & Lively & 19 & Artificial & Lifelike  \\ 
8 & Unconscious & Conscious  & 20 & Ignorant & Knowledgeable  \\
9 & Mechanical & Organic & 21 & Anxious & Relaxed  \\ 
10 & Apathetic & Responsive & 22 & Calm  & Agitated \\
11 & Awful & Nice & 23 & Quiescent &  Surprised \\ 
12 &  Machinelike & Humanlike &    &   &   \\ 
\hline
\hline
\end{tabular}
}
\label{tbl:SD}
\end{table}

\begin{table}
\tbl{Questions of ranking method}{
\begin{tabular}{cl}
\hline
\hline

No. & \multicolumn{1}{c} {Question} \\
\hline
1 & In which experiment did you feel affinity for the robot? \\
2 & In which experiment did you feel the intelligence of the robot? \\
3 & In which experiment did you feel a sense of security? \\
4 & In which experiment did you feel friendliness or ease of talking to the remote operator?\\
\hline
\hline
\end{tabular}
}
\label{tbl:Q_Order}
\end{table}

\section{Results}
\label{sec:result}

\subsection{Results of SD Method}

\subsubsection{Confirmation of Influence of Subject Demographics and Test Order}
\label{sec:BGandTO}

\begin{figure}
\begin{center}
\includegraphics[width=100mm]{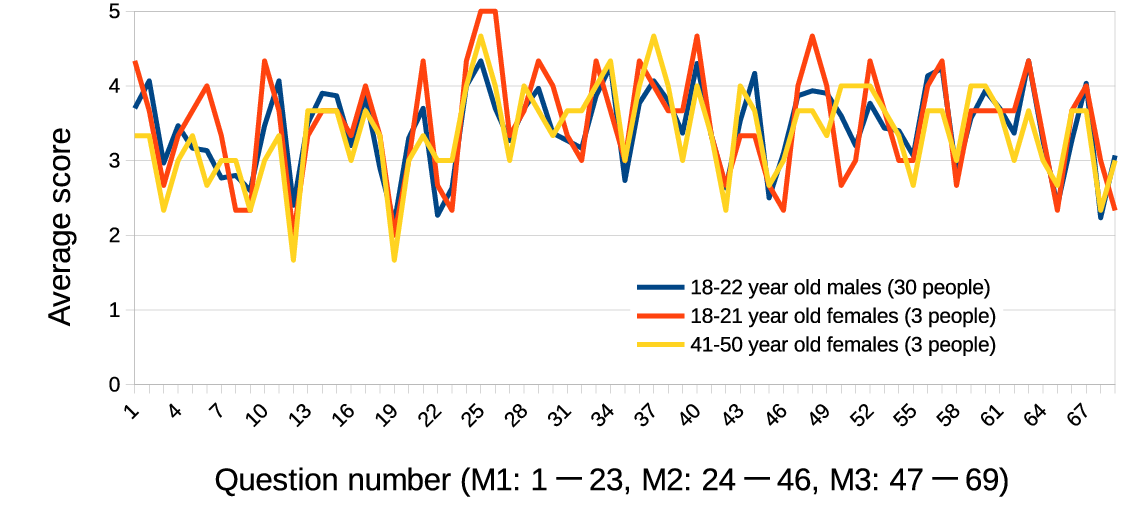}
\caption{Average SD method scores for 3 subject classes}
\label{fig:M123MW}
\end{center}
\end{figure}

First, we confirmed the effects of subject demographics on affinity. We divided all the subjects into three classes, namely 18-22 year old males (30 engineering course students), 18-21 year old females (3 engineering course students), and 41-50 year old females (3 university officers). The averages of the questions and answers for each class are presented in Figure~\ref{fig:M123MW}. Here, the horizontal axis is the question number for each control mode, where Q1 to Q23 are for M1, Q24 to Q46 are for M2, and Q47 to Q69 are for M3. Because the number of data on the latter two classes are limited, they cannot be analyzed statistically. However, it can be observed that the averages tend to be the same, even if the demographics differed. Hence, we used all the data equally to conduct the next step. Accurate statistical analysis of the effects of differences in subject demographics could be a topic for further study.

\begin{figure}
\begin{center}
\includegraphics[width=100mm]{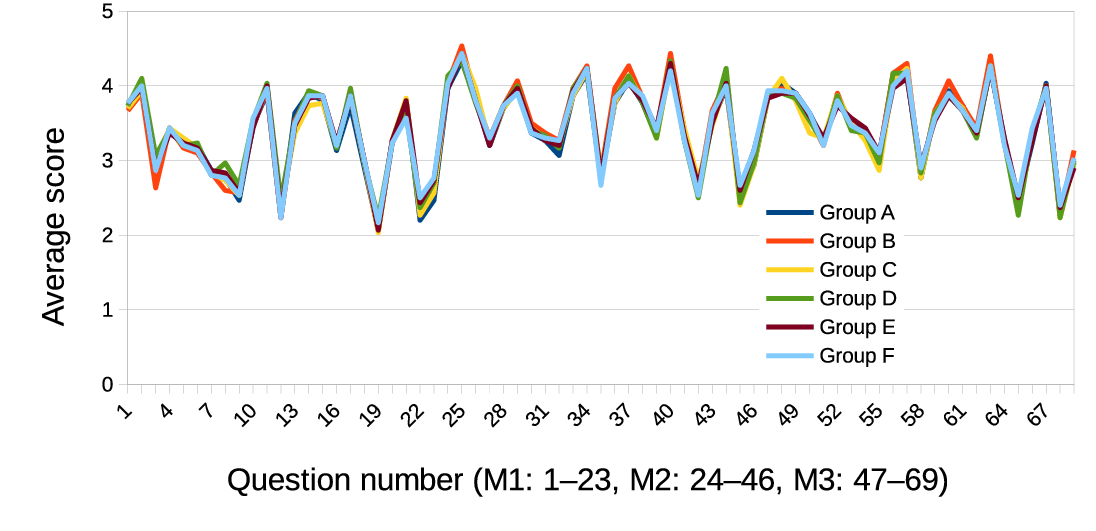}
\caption{Average SD method scores for subject Groups A--F}
\label{fig:M123}
\end{center}
\end{figure}

As described in Section~\ref{sec:abs_exp}, we established six groups of subjects (Groups A--F) with different orders of the control modes. Prior to the analysis, we confirmed the effects of the difference among the groups on the answer. The averages of the questions and answers for each group are presented in Figure~\ref{fig:M123}. It can be observed that all the averages were around the same, even if the subject groups differed.

\begin{figure}
\begin{center}
\includegraphics[width=100mm]{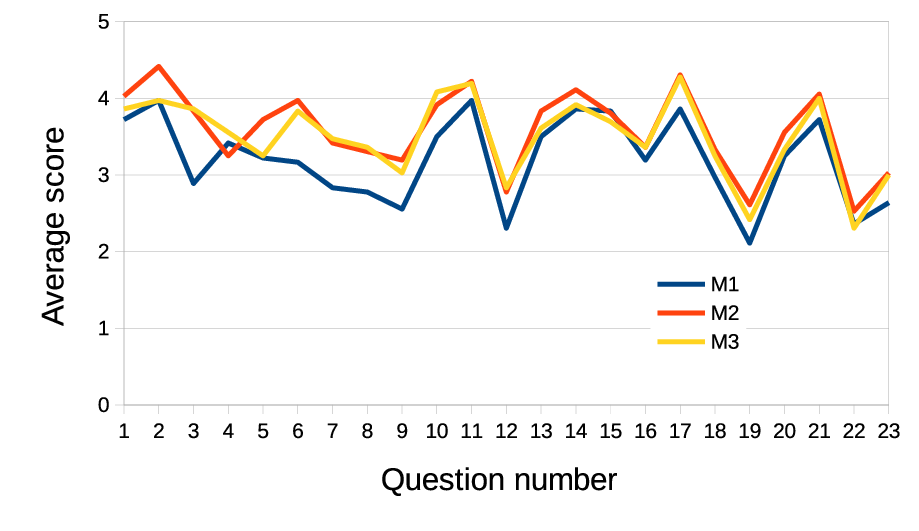}
\caption{Average values for each question of SD method}
\label{fig:SD_FULL}

\end{center}
\end{figure}

Furthermore, when one-way analysis of variance was performed for each question, no significant difference $ (p <0.01) $ was identified for any question regarding the differences among the population means of the six subjects. According to these results, it was decided to carry out the analysis without considering the influence of the test order. A graph of the average values of the 36 subjects in each control mode (M1 to M3) for each question of the SD method, without considering the test order, is presented in Figure~\ref{fig:SD_FULL}, and the values of the mean and standard error (SE) are displayed in Table~\ref{tbl:SD_FULL}.

\subsubsection{Analysis of Single Question}

\begin{figure}
\begin{center}
\includegraphics[width=100mm]{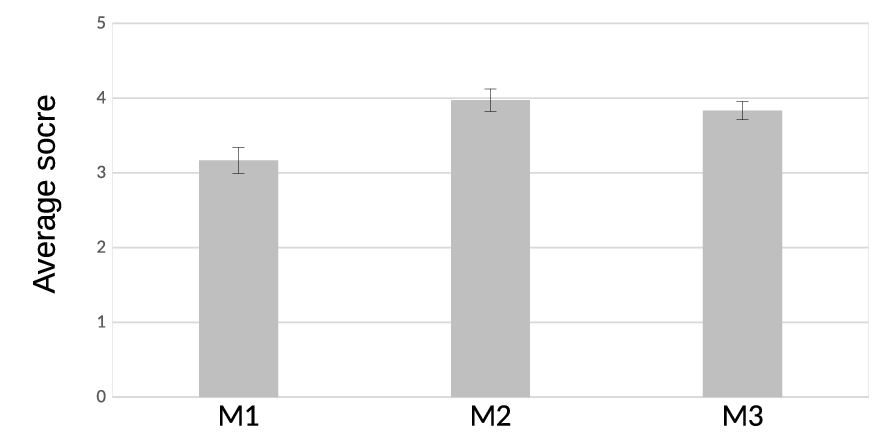}
\caption{Average score and SE of question no. 6}
\label{fig:M_1word}
\end{center}
\end{figure}

As the first step of the analysis, we analyzed Q6 (`Unfriendly--Friendly'), which was one of the most direct adjective pairs in this study. Figure~\ref{fig:M_1word} illustrates the average score (bar graph) and standard error (SE, error bar) of Q6 for each control mode.

Next, Holm's multiple comparison method was performed on the $t$-test for the difference among the population means in each control mode. As indicated in Table~\ref{tbl:shitashimi}, the population means of M2 and M3 were significantly higher than that of M1 ($p < 0.01$). For the effect size of Cohen's $d$, the difference between M2 and M1 was quite large ($d = 0.829 > 0.8$), and the difference between M3 and M1 was medium to large ($0.5 < d = 0.742 < 0.8$). According to these results, the affinities of M2 and M3 were higher than that of M1, and the difference in the effect sizes indicates that the affinity of M2 was higher than that of M3.
\begin{table}
\tbl{Results of multiple comparison method and effect size of question no. 6}
{
\begin{tabular}{c|cc}
\hline
\hline
Comparison & $p$ &  $d$  \\
\hline
M2 > M1 &  $0.003^{**}$ & 0.829  \\
M3 > M1 &   $0.002^{**}$ & 0.742 \\
M2 > M3 & $0.405^{\ \  } $ & 0.170 \\
\hline
\hline
\end{tabular}
}
\tabnote{$\dag: p < 0.1, *: p < 0.05, **: p < 0.01$}
\label{tbl:shitashimi}
\end{table}

\subsubsection{Factor Analysis}

\begin{figure}
\begin{center}
\includegraphics[width=10cm]{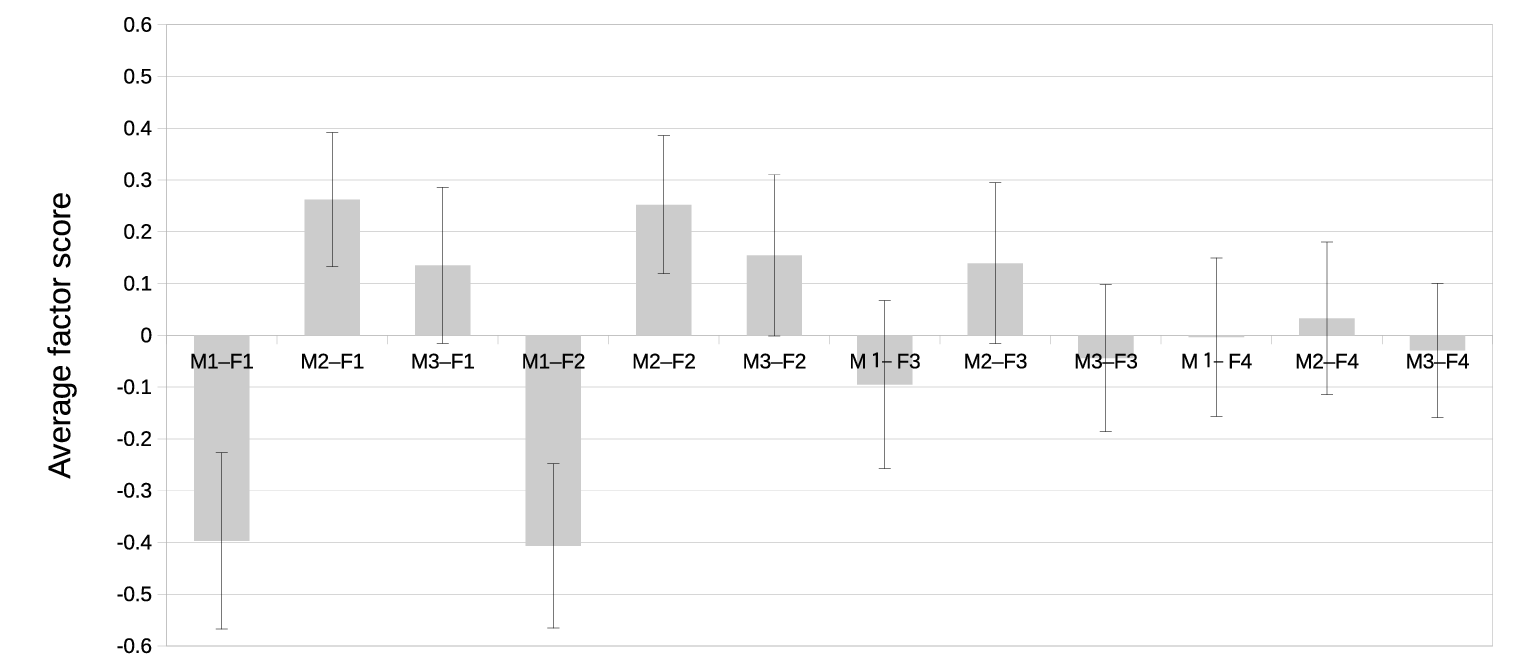}
\caption{Average factor score and SE in control modes M1 to M3 and factors F1 to F4}
\label{fig:FactoScore}
\end{center}
\end{figure}

We used the stringing-out method for all the data from the three control modes performed by the 36 subjects. Therefore, the total number of observation data was $108(= 36 \times 3)$. The observed variables were the questions of the SD methods (Table~\ref{tbl:SD}), and the number of observed variables was 23. We used the Kaiser--Guttman criterion to determine the number of factors. As the eigenvalues of the correlation matrix were $9.07, 1.89, 1.59, 1.35, 0.95, \dots$ in order of magnitude, four factors were determined with eigenvalues of 1.00 or more. Table~\ref{tab:FA} presents the results of the factor analysis with factor extraction and rotation methods used as maximum likelihood and promax methods, respectively. We used the statistical computing software R version 3.4.4 for the calculations. From the observed variables  Table~\ref{tab:FA}, we defined the first factor as vitality, the second as affinity, the third as intellect, and the fourth as inciting. The graph of the averages and SEs of the factor scores are displayed in Figure~\ref{fig:FactoScore}. The statistics are presented in Table~\ref{tbl:FA2}, and results of the multiple comparison method ($t$-test, Holm's method) and effect sizes (Cohen's $d$) are provided in Table~\ref{tbl:FA3}. We excluded the factor F4 from the multiple comparison analysis, because the maximum mean difference among the control modes was extremely small at $0.06 (= 0.03-(-0.03))$ in Table~\ref{tbl:FA2}, and the factor correlation matrix in Table~\ref{tab:FA} indicates low correlation with the other factors and high independence. The affinity (F2) was significantly higher in M2 and M3 than in M1 ($p < 0.01$). The effect sizes of M2 and M3 from M1 were sufficiently large ($d > 0.8$) and medium to large ($0.5 < d < 0.8$), respectively. According to these results, the affinities of M2 and M3 were improved from M1. Although the effect size of M2 from M3 was slightly positive ($0 < d < 0.2$), it could be concluded that the affinity of M2 was higher than that of M3. A similar tendency was observed for the vitality (F1). For the intellect (F3), the comparison was not statistically significant in any of the multiple comparison methods. However, although the effect sizes exhibited small values, they were in the same order as vitality and affinity.
\begin{table}
\tbl{Factor analysis results}
{
\begin{tiny} 
\begin{tabular} {lrrrrr }
 \multicolumn{ 5 }{l}{ Factor loading matrix   } \cr 
 \hline 
 Observed variable &   F1 &  F2  & F3  &  F4  &   h2 \cr 
 (right adjective) &  Vitality &  Affinity  &  Intellect  & Inciting &  Commonality \cr 
  \hline 
 Humanlike   &  \bf{ 0.87}  &  -0.08  &  -0.09  &   0.09  &  0.62 \cr 
 Lifelike   &  \bf{ 0.81}   &  -0.12  &   0.08  &   0.09  &  0.62   \cr 
 Organic   &  \bf{ 0.78}  &   0.03  &   0.02  &   0.05  &  0.66 \cr 
 Conscious   &  \bf{ 0.67}  &   0.03  &   0.09  &   0.16  &  0.59  \cr 
 Alive   &  \bf{ 0.54}  &   0.23  &   0.07  &  -0.05   &  0.58   \cr 
 Lively   &  \bf{ 0.51}  &  0.33  &  -0.11  &  -0.10  &  0.50\cr 
 Interactive   &  \bf{ 0.48}  &   0.22  &  -0.05  &  -0.09  &  0.39 \cr 
 Natural   &  \bf{ 0.47}  &   0.12  &   0.05  &   0.05  &  0.36 \cr 
 Like   &  -0.01  &  \bf{ 0.90}  &  -0.09  &   0.09  &  0.69 \cr 
 Friendly  &   0.15  &  \bf{ 0.65}  &  -0.14  &  -0.09  &  0.47 \cr 
 Nice   &  -0.26  &  \bf{ 0.64}  &  0.44  &   0.04  &  0.66 \cr 
 Pleasant   &   0.01  &  \bf{ 0.57}  &   0.02  &   0.19  &  0.37 \cr 
 Relaxed  &   0.08  &  \bf{ 0.47}  &  -0.07  &  {-0.48}  &  0.50   \cr 
 Kind   &   0.22  &  \bf{ 0.42}  &   0.13  &  -0.01  &  0.46 \cr 
 Responsive   &   0.15  &  \bf{ 0.43}  &   0.24  &   0.09  &  0.53  \cr 
 Responsible   &   0.14  &  \bf{ 0.32}  &   0.27  &  -0.10  &  0.42 \cr 
 Sensible   &  -0.06  &  -0.10  &  \bf{ 0.88}  &   0.05  &  0.63  \cr 
 Moving elegantly   &   0.04  &  -0.08  &  \bf{ 0.63}  &  -0.12  &  0.38 \cr 
 Competent   &  -0.06  &  { 0.31}  &  \bf{ 0.54}  &   0.03  &  0.54 \cr 
 Intelligent   &  { 0.40}  &  -0.20  &  \bf{ 0.56}  &  -0.08  &  0.53 \cr 
 Knowledgeable  &   0.07  &   0.19  &  \bf{ 0.51}  &   0.02  &  0.49 \cr 
 Surprised  &  -0.10  &  { 0.43}  &  -0.03  &  \bf{ 0.73}  &  0.62 \cr 
 Agitated &   0.28  &  -0.11  &  -0.08  &  \bf{ 0.67}  &  0.49 \cr 

\hline \cr 
 Cumulative percentage & 0.35  & 0.66 & 0.89 & 1.00 &  \cr  
\cr 

 \multicolumn{ 5 }{l}{ Factor correlation matrix } \cr 
 \hline 
      &  F1 &  F2 &  F3 &  F4 \cr 
  \hline       
F1   &  1.00 &  0.66 &  0.60 &  0.00 \cr 
F2   &  0.66 &  1.00 &  0.64 & -0.07 \cr 
F3   &  0.60 &  0.64 &  1.00 &  0.02 \cr 
F4   &  0.00 & -0.07 &  0.02 &  1.00 \cr 
 \hline 

\end{tabular}
\end{tiny}
}
\label{tab:FA}
\end{table} 

\begin{table}
\tbl{Factor score statistics}
{
\begin{tabular}{l|cc|cc|cc}
\hline
\hline
 & \multicolumn{2}{c}{M1} &  \multicolumn{2}{c}{M2}&  \multicolumn{2}{c}{M3}\\
 \hline
 Factor & Mean & SE & Mean & SE & Mean & SE\\ 
 \hline
F1  & -0.40 & 0.17 & 0.26 & 0.13 & 0.14 & 0.15 \\ 
F2 & -0.41 & 0.16 & 0.25 & 0.13 & 0.15 & 0.16 \\ 
F3 & -0.10 & 0.16 & 0.14 & 0.15 & -0.04 & 0.14 \\ 
F4 & -0.00 & 0.15 & 0.03 & 0.15 & -0.03 & 0.13 \\ 

\hline
\hline
\end{tabular}
}
\label{tbl:FA2}
\end{table}

\begin{table}
\tbl{Results of multiple comparison method of factor scores and effect sizes}
{
\begin{tabular}{l|cc|cc|cc}
\hline
\hline
& \multicolumn{2}{c}{F1 Vitality} &  \multicolumn{2}{c}{F2 Affinity}&  \multicolumn{2}{c}{F3 Intellect}\\
Comparison & $p$  &  $d$    & $p$  &  $d$  & $p$  &  $d$    \\
\hline
   M2 > M1     & $0.004^{**}$ & 0.773   &  ${0.000}^{**}$ & 0.840 &  0.204 & 0.257 \\
   M3 > M1 & $0.006^{**}$  & 0.567   &  ${0.000}^{**}$ & 0.631  &  1.346 & 0.060\\
  
 M2 > M3 & 0.425  & 0.176   &  $0.411$ & 0.128  &  0.561 & 0.228 \\

\hline
\hline
\end{tabular}
}
\tabnote{$\dag: p < 0.1, *: p < 0.05, **: p < 0.01$}
\label{tbl:FA3}
\end{table}

\subsection{Results of ranking method}

\begin{figure}
\begin{center}
\includegraphics[width=10cm]{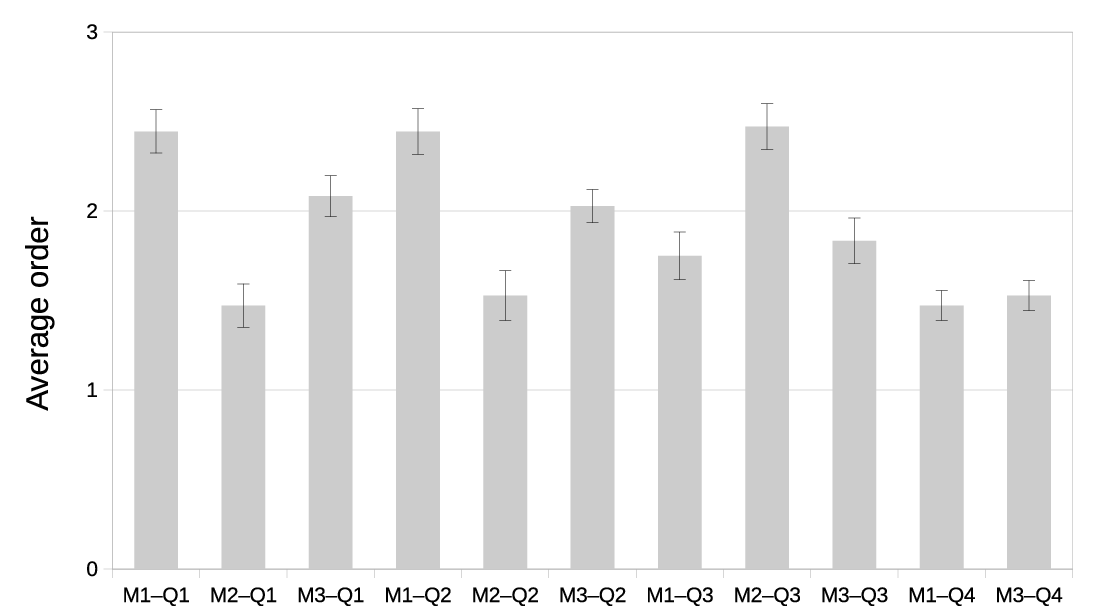}
\caption{Average order and SE in control modes M1 to M3 and questions Q1 to Q4}
\label{fig:AverageOrder}
\end{center}
\end{figure}

Figure~\ref{fig:AverageOrder} and Table~\ref{tbl:Order} present the results of the ranking method. The figure and table display the average values of the ranks of the answers of the 36 subjects (first, second, and third), where smaller values indicate more positive answers. The results of the multiple comparison method are presented in Table~\ref{tbl:Order2}. The $p$ values were obtained from Wilcoxon's signed-rank test and modified by Holm's method. The effect size $d$ value was calculated using Cliff's $d$. Conversely, a larger $d$ indicates a higher rank and smaller rank value. Regarding the average order and effect size, the affinity for $R$ (Q1) was positive in the order of M2, M3, and M1, as was the case in the SD method. Regarding the intellect (Q2), the same tendency as the affinity was observed. However, for the sense of security (Q3), the order was reversed, unlike in Q1 and Q2. That is, it was found that M1 and M3 were more secure than M2 from the perspective of the effect size ($d = -0.500, -0.389$). The effect size was extremely small between M1 and M3 ($d = 0.000$), and no significant difference was observed. Regarding the affinity for remote operator $C$ (Q4), no significant difference was found from the average ranks of M1 and M3, which had almost the same means at 1.47 and 1.53, respectively, as indicated in Table~\ref{tbl:Order}.
\begin{table}
\tbl{Ranking method statistics in each control mode}
{
\begin{tabular}{l|cc|cc|cc}
\hline
\hline
 & \multicolumn{2}{c}{M1} &  \multicolumn{2}{c}{M2}&  \multicolumn{2}{c}{M3}\\
 \hline
 Factor & Mean & SE & Mean & SE & Mean & SE\\ 
 \hline
Q1 Affinity for $R$ & 2.44 & 0.12 & 1.47 & 0.12 & 2.08 & 0.11 \\ 
Q2 Intellect & 2.44 & 0.13 & 1.53 & 0.14 & 2.03 & 0.09 \\ 
Q3 Sense of security& 1.75& 0.13& 2.47 & 0.13 & 1.83 & 0.13 \\ 
Q4 Affinity for $C$ &1.47  & 0.08 & - &  -& 1.53 & 0.08 \\ 
\hline
\hline
\end{tabular}
}
\label{tbl:Order}
\end{table}

\begin{table}
\tbl{Results of multiple comparison method for population means of ranking method and effect size}
{
\begin{tabular}{l|cc|cc|cc}
\hline
\hline
Comparison& \multicolumn{2}{c}{Affinity} &  \multicolumn{2}{c}{Intellect}&  \multicolumn{2}{c}{Sense of security}\\
 & $p$  &  $d$    & $p$  &  $d$  & $p$  &  $d$    \\
\hline
   M2 > M1     & $0.001^{**}$ & 0.611 & $0.005^{**}$  & 0.500 &  $0.017^{*}$ & -0.500 \\
   M3 > M1 & $0.099^{\dag}$ & 0.278   &  $0.068^{\dag}$ & 0.389  &  0.753  & 0.000\\
   M2 > M3 & $0.015^{*}  $ & 0.444   &  $0.026^{*} $       & 0.444   &  $0.013^{*}$ & -0.389 \\
\hline
\hline
\end{tabular}
}
\tabnote{$\dag: p < 0.1, *: p < 0.05, **: p < 0.01$}
\label{tbl:Order2}
\end{table}

\section{Discussion}
\label{sec:discussion}

The results in the previous section demonstrate that the affinity of the robot exhibit the order: autonomous control mode, autonomous remote control mode, and remote control mode in both SD and ranking method. The same tendencies of affinity were exhibited in the vitality and intellect when using SD, and ranking method. However, the ranking method indicate that the remote control and autonomous remote control modes had higher sense of security than the autonomous control mode. In the SD method, although there was a direct question about the sense of security (no. 21, `Anxious--Relaxed'), the means of M2 and M3 were higher than that of M1, which resulted from the factor analysis being incorporated into the affinity (F2). Therefore, the results of the SD method differed from those of the ranking method. Q4 of the ranking method was meant to examine the impression of the remote operator according to the difference between two-party and three-party system. The results demonstrated no significant difference between the two.
 
According to these results, if we assume that remote functions are still required at present to execute various physical tasks, we can conclude that it is effective to introduce the robot agent explicitly to the user as the autonomous remote control robot, from the perspective of improving the affinity. The dialogue used in this study is a simple command dialogue that is required for physical task instructions, and it is believed that the requested autonomy for the autonomous remote control mode will be achievable at present. Regarding the sense of security, several subjects might have reconsidered the contribution of the operator agent once the experiments were over. Considering that autonomous robot technology remains under development and has not yet been widely used, it is understandable that the subjects desired the sense of security that is constantly observed by humans remotely. There was no significant difference between the impressions of the remote operator in the two control modes in Q4 of the ranking method, but we believe that it is too early to decide that the question has no significance. Although the evaluation of autonomous functions to make the operations easier for remote operators has been researched extensively, the impression of remote operators with different robot configurations has not been substantially evaluated. By analogy with human relationships, it may be more efficient to work while consulting with three people. If the task progresses through the exchange of only two people, apart from the user, the user may experience negative feelings such as alienation. Although it is an ambiguous and complicated problem, the impression of the remote operator is also important for the social implementation of the autonomous remote control robot system, including remote support services. Hence, we consider this as an important future research topic.

\section{Conclusions}
\label{sec:conclusion}

We conclude that the affinity of autonomous control robots is higher than that of remote control robots, and less than that of autonomous robots, as assumed in Section~\ref{sec:intro}, based on our studies.
First, we defined the autonomous remote control robot as a robot performing physical tasks with the robot and operator agents and compared its affinity with those of remote control and autonomous robots. For comparison, we selected an FCT as the most basic task. We defined the task conditions and clarified the scope of this study. The research platform HSR was adopted to conduct the experiments, and the experimental environment was implemented per the WRS 2020 Rulebook, with consideration for future comparisons and additional evaluations. Using the results of the subject experiments, we performed analyses with the SD and ranking method. In each case, higher autonomy resulted in higher affinity. Moreover, similar results were obtained for factors such as vitality and intellect. According to the results, we believe that autonomous remote control robots provide a useful research direction for robots that perform physical tasks in environments where human beings coexist. With such a robot, it is possible to cope with various tasks by remote functions and to improve the autonomous functions sequentially along with its affinity.

\section{Acknowledgment}
We would like to thank Dr. Takuto Sakuma and Dr. Luis Contreras-Toledo for their helpful advice in conducting this study.

\bibliography{hsrAR}

@article{riek2012wizard,
  title={Wizard of oz studies in hri: a systematic review and new reporting guidelines},
  author={Riek, Laurel D},
  journal={Journal of Human-Robot Interaction},
  volume={1},
  number={1},
  pages={119--136},
  year={2012},
  publisher={Journal of Human-Robot Interaction Steering Committee}
}

@inproceedings{bohren2011towards,
  title={Towards autonomous robotic butlers: Lessons learned with the PR2},
  author={Bohren, Jonathan and Rusu, Radu Bogdan and Jones, E Gil and Marder-Eppstein, Eitan and Pantofaru, Caroline and Wise, Melonee and M{\"o}senlechner, Lorenz and Meeussen, Wim and Holzer, Stefan},
  booktitle={2011 IEEE International Conference on Robotics and Automation},
  pages={5568--5575},
  year={2011},
  organization={IEEE}
}

@inproceedings{yamaguchi2019live,
  title={Live Demonstration: A VLSI Implementation of Time-Domain Analog Weighted-Sum Calculation Model for Intelligent Processing on Robots},
  author={Yamaguchi, Masatoshi and Iwamoto, Gouki and Abe, Yushi and Tanaka, Yuichiro and Ishida, Yutaro and Tamukoh, Hakaru and Morie, Takashi},
  booktitle={2019 IEEE International Symposium on Circuits and Systems (ISCAS)},
  pages={1--1},
  year={2019},
  organization={IEEE}
}

@inproceedings{yamamoto2019human,
  title={Human Support Robot as Research Platform of Domestic Mobile Manipulator},
  author={Yamamoto, Takashi and Takagi, Yutaro and Ochiai, Akiyoshi and Iwamoto, Kunihiro and Itozawa, Yuta and Asahara, Yoshiaki and Yokochi, Yasukata and Ikeda, Koichi},
  booktitle={Robot World Cup},
  pages={457--465},
  year={2019},
  organization={Springer}
}

@inproceedings{berenstein2018robustly,
  title={Robustly adjusting indoor drip irrigation emitters with the toyota hsr robot},
  author={Berenstein, Ron and Fox, Roy and McKinley, Stephen and Carpin, Stefano and Goldberg, Ken},
  booktitle={2018 IEEE International Conference on Robotics and Automation (ICRA)},
  pages={2236--2243},
  year={2018},
  organization={IEEE}
}

@inproceedings{yi2019mobile,
  title={Mobile Manipulation for the HSR Intelligent Home Service Robot},
  author={Yi, Jae-Bong and Yi, Seung-Joon},
  booktitle={2019 16th International Conference on Ubiquitous Robots (UR)},
  pages={169--173},
  year={2019},
  organization={IEEE}
}

@inproceedings{pena2018eeva,
  title={eEVA as a Real-Time Multimodal Agent Human-Robot Interface},
  author={Pe{\~n}a, P and Polceanu, M and Lisetti, C and Visser, U},
  booktitle={Robot World Cup},
  pages={262--274},
  year={2018},
  organization={Springer}
}

@inproceedings{quispe2017learning,
  title={Learning user preferences for robot-human handovers},
  author={Quispe, Ana C Huam{\'a}n and Martinson, Eric and Oguchi, Kentaro},
  booktitle={2017 IEEE/RSJ International Conference on Intelligent Robots and Systems (IROS)},
  pages={834--839},
  year={2017},
  organization={IEEE}
}

@inproceedings{itadera2019impedance,
  title={Impedance Control based Assistive Mobility Aid through Online Classification of User’s State},
  author={Itadera, Shunki and Kobayashi, Taisuke and Nakanishi, Jun and Aoyama, Tadayoshi and Hasegawa, Yasuhisa},
  booktitle={2019 IEEE/SICE International Symposium on System Integration (SII)},
  pages={243--248},
  year={2019},
  organization={IEEE}
}

@article{duffy2003anthropomorphism,
  title={Anthropomorphism and the social robot},
  author={Duffy, Brian R},
  journal={Robotics and autonomous systems},
  volume={42},
  number={3-4},
  pages={177--190},
  year={2003},
  publisher={Elsevier}
}

@article{heerink2008influence,
  title={The influence of social presence on acceptance of a companion robot by older people},
  author={Heerink, Marcel and Kr{\"o}se, Ben and Evers, Vanessa and Wielinga, Bob},
  year={2008},
  publisher={Red de Agentes F{\'\i}sicos}
}

@article{gaudiello2016trust,
  title={Trust as indicator of robot functional and social acceptance. An experimental study on user conformation to iCub answers},
  author={Gaudiello, Ilaria and Zibetti, Elisabetta and Lefort, S{\'e}bastien and Chetouani, Mohamed and Ivaldi, Serena},
  journal={Computers in Human Behavior},
  volume={61},
  pages={633--655},
  year={2016},
  publisher={Elsevier}
}

@inproceedings{klamer2010acceptance,
  title={Acceptance and use of a social robot by elderly users in a domestic environment},
  author={Klamer, Tineke and Allouch, Somaya Ben},
  booktitle={2010 4th International Conference on Pervasive Computing Technologies for Healthcare},
  pages={1--8},
  year={2010},
  organization={IEEE}
}

@inproceedings{ogawa2008itaco,
  title={ITACO: Constructing an emotional relationship between human and robot},
  author={Ogawa, Kohei and Ono, Tetsuo},
  booktitle={RO-MAN 2008-The 17th IEEE International Symposium on Robot and Human Interactive Communication},
  pages={35--40},
  year={2008},
  organization={IEEE}
}

@inproceedings{hashimoto2013field,
  title={A field study of the human support robot in the home environment},
  author={Hashimoto, Kunimatsu and Saito, Fuminori and Yamamoto, Takashi and Ikeda, Koichi},
  booktitle={2013 IEEE Workshop on Advanced Robotics and its Social Impacts},
  pages={143--150},
  year={2013},
  organization={IEEE}
}

@inproceedings{nagahama2018learning,
  title={A Learning Method for a Daily Assistive Robot for Opening and Closing Doors Based on Simple Instructions},
  author={Nagahama, Kotaro and Takeshita, Keisuke and Yaguchi, Hiroaki and Yamazaki, Kimitoshi and Yamamoto, Takashi and Inaba, Masayuki},
  booktitle={2018 IEEE 14th International Conference on Automation Science and Engineering (CASE)},
  pages={599--605},
  year={2018},
  organization={IEEE}
}

@inproceedings{ando2018experimental,
  title={Experimental Evaluation of Haptic Visualization Interface for Robot Teleoperation Using Onomatopoeia in a Haptic Recognition Task},
  author={Ando, Maika and Chiba, Jotaro and Itadera, Shunki and Nakanishi, Jun and Aoyama, Tadayoshi and Hasegawa, Yasuhisa},
  booktitle={2018 International Symposium on Micro-NanoMechatronics and Human Science (MHS)},
  pages={1--4},
  year={2018},
  organization={IEEE}
}

@incollection{tachi1985tele,
  title={Tele-existence (I): Design and evaluation of a visual display with sensation of presence},
  author={Tachi, Susumu and Tanie, Kazuo and Komoriya, Kiyoshi and Kaneko, Makoto},
  booktitle={Theory and Practice of Robots and Manipulators},
  pages={245--254},
  year={1985},
  publisher={Springer}
}

@article{kanda2001e,
  title={Psychological Evaluation on Interactions between People and Robot},
  author={Takayuki Kanda and Hiroshi Ishiguro and Toru Ishida},
  journal={Journal of the Robotics Society of Japan},
  volume={19},
  number={3},
  pages={362--371},
  year={2001},
  publisher={he Robotics Society of Japan}
}

@article{yamaoka2007e,
  title={Interacting with a Human or a Humanoid Robot?},
  author={Fumitaka Yamaoka and Takayuki Kanda and Hiroshi Ishiguro and Norihiro Hagita},
  journal={Transactions of Information Processing Society of Japan},
  volume={48},
  number={11},
  pages={3577--3587},
  year={2007}
}

@inproceedings{2011touchmeE,
  title={TouchMe: An Augmented Reality Based Remote Robot Manipulation},
  author={Sunao Hashimoto and Akihiko Ishida and Masahiko Inami and Takeo Igarashi},
  booktitle={The 21st International Conference on Artificial Reality and Telexistence, Proceedings of ICAT},
  year={2011}
}

@inproceedings{2009sketch,
  title={Sketch and run: a stroke-based interface for home robots},
  author={Sakamoto, Daisuke and Honda, Koichiro and Inami, Masahiko and Igarashi, Takeo},
  booktitle={Proceedings of the SIGCHI Conference on Human Factors in Computing Systems},
  pages={197--200},
  year={2009},
  organization={ACM}
}

@book{Okubo2012,
  title={Psychological statistics to convey: effect size, confidence interval, and test power},
  author={Matia Okubo and Kensuke Okada},
  year={2012},
  publisher={Keiso Shobo}
}

@article{bartneck2009measurement,
  title={Measurement instruments for the anthropomorphism, animacy, likeability, perceived intelligence, and perceived safety of robots},
  author={Bartneck, Christoph and Kuli{\'c}, Dana and Croft, Elizabeth and Zoghbi, Susana},
  journal={International journal of social robotics},
  volume={1},
  number={1},
  pages={71--81},
  year={2009},
  publisher={Springer}
}

@article{Kanda2002,
  title={Effects of Observation of Robot-Robot Communication on Human-Robot Communication},
  author={Takayuki Kanda and Hiroshi Ishiguro and Tetsuo Ono and Michita Imai and Ryohei Nakatsu},
  journal={The IEICE Transactions on Information and Systems(Japanese Edition) },
  volume={85},
  number={7},
  pages={691--700},
  year={2002},
  publisher={The Institute of Electronics, Information and Communication Engineers}
}

@article{Fujie2012,
  title={Conversation Robot Participating in and Promoting Human-Human Communication},
  author={Shinya Fujie and Yoichi Matsuyama and Hikaru Taniyama and Tetsunori Kobayashi},
  journal={The Transactions of The Institute of Electronics, Information and Communication Engineering (IEICE) A},
  volume={95},
  number={1},
  pages={37--45},
  year={2012},
  publisher={The Institute of Electronics, Information and Communication Engineers}
}

@article{Matsusaka2001,
  title={Conversation Robot Participating in Group Conversation},
  author={Yosuke Matsusaka and Tsuyoshi Tojo and Tetsunori Kobayashi},
  journal={The IEICE Transactions on Information and Systems(Japanese Edition)},
  volume={84},
  number={6},
  pages={898--908},
  year={2001},
  publisher={The Institute of Electronics, Information and Communication Engineers}
}

@inproceedings{kato2004development,
  title={Development of a communication robot ifbot},
  author={Kato, Shohei and Ohshiro, Shingo and Itoh, Hidenori and Kimura, Kenji},
  booktitle={IEEE International Conference on Robotics and Automation, 2004. Proceedings. ICRA'04. 2004},
  volume={1},
  pages={697--702},
  year={2004},
  organization={IEEE}
}

@inproceedings{iwata2018learning,
  title={Learning and Generation of Actions from Teleoperation for Domestic Service Robots},
  author={Iwata, Kensuke and Aoki, Tatsuya and Horii, Takato and Nakamura, Tomoaki and Nagai, Takayuki},
  booktitle={2018 IEEE/RSJ International Conference on Intelligent Robots and Systems (IROS)},
  pages={8184--8191},
  year={2018},
  organization={IEEE}
}

@article{okada2019competitions,
  title={What competitions were conducted in the service categories of the World Robot Summit?},
  author={Okada, Hiroyuki and Inamura, Tetsunari and Wada, Kazuyoshi},
  journal={Advanced Robotics},
  volume={33},
  number={17},
  pages={900--910},
  year={2019},
  publisher={Taylor \& Francis}
}

@inproceedings{borst2009rollin,
  title={Rollin'justin-mobile platform with variable base},
  author={Borst, Christoph and Wimbock, Thomas and Schmidt, Florian and Fuchs, Matthias and Brunner, Bernhard and Zacharias, Franziska and Giordano, Paolo Robuffo and Konietschke, Rainer and Sepp, Wolfgang and Fuchs, Stefan and others},
  booktitle={2009 IEEE International Conference on Robotics and Automation},
  pages={1597--1598},
  year={2009},
  organization={IEEE}
}

@article{watanabe2013cooking,
  title={Cooking behavior with handling general cooking tools based on a system integration for a life-sized humanoid robot},
  author={Watanabe, Yoshiaki and Nagahama, Kotaro and Yamazaki, Kimitoshi and Okada, Kei and Inaba, Masayuki},
  journal={Paladyn, Journal of Behavioral Robotics},
  volume={4},
  number={2},
  pages={63--72},
  year={2013},
  publisher={Versita}
}

@article{Robomech2019,
title={Development of Human Support Robot as the research platform of a domestic mobile manipulator},
  author={Yamamoto, Takashi and Terada, Koji and Ochiai, Akiyoshi and Saito, Fuminori and Asahara, Yoshiaki and Murase, Kazuto},
  journal={ROBOMECH Journal},
  volume={6},
  number={1},
  pages={4},
  year={2019},
  publisher={Springer}
}

@electronic{HSR,
organization = "TOYOTA MOTOR CORPORATION",
title = {{Toyota Shifts Home Helper Robot R\&D into High Gear with
New Developer Community and Upgraded Prototype}},
url = "https://newsroom.toyota.co.jp/en/detail/8709541",
month = "July 16",
year = "2015"
}
\bibliographystyle{unsrt}

\appendices
\label{sec:appendix}

\section{Examples of Dialogue in Three Control Modes}
\label{sec:appendix_dialog}

\subsection{Example of  Dialogue in Remote Control Mode (M1)}
\label{sec:appendix_dialog1}
\par\noindent
$C$: Hello, $U$.  \\
$U$: Hello.\\
$C$: I am Helper $C$. Nice to meet you. $U$, is there anything you want?\\ 
$U$: I want a pen. \\
$C$: All right.  Wait a moment please. ($R$ moves and arrives at the table.) \\
$C$: $U$, which pen do you want?\\
$U$: I want the red pen in the front. \\
$C$: I understand. (Show the line drawn on the object by $C$ on $U$'s PC screen.) \\
$C$: I have drawn a red line on the screen. Is this right? \\
$U$: Yes. OK. \\
$C$: I will take it. \\
$C$:  $U$, is this OK? (The object in $R$'s hand is displayed on $U$'s PC screen.) \\
$U$: OK. \\
$C$: Thank you. I will bring it to you. ($R$ moves and arrives at $U$.) \\
$C$: $U$, here you are. ($U$ receives object $O$.) \\
$U$: Thank you. \\
$C$: You are welcome.\\

\subsection{Example of Dialogue in Autonomous Control Mode (M2)}
\label{sec:appendix_dialog2}
\par\noindent
$R$: Hello.\\
$U$: Hello.\\
$R$: Is there anything you want, $U$?\\
$U$: $R$, I want a pen.\\
$R$: All right. The pen is on the table, so I will move there.\\
$U$: OK.\\
$R$: Now, I am in front of the table. Which pen do you want? \\
$U$: $R$, I want the red pen in the front. \\
$R$: All right. I will take it. Wait a moment please.\\
$R$: Done. $U$, is this OK? (The object in $R$'s hand is displayed on $U$'s PC screen.) \\
$U$: $R$, it is OK.\\
$R$: All right. I will bring it to you. ($R$ moves and arrives at $U$.)\\
$R$: Here you are.\\
$U$: $R$, thank you.\\
$R$: You are welcome.\\

\subsection{Example of Dialogue in Autonomous Remote Control Mode (M3)}
\label{sec:appendix_dialog3}
\par\noindent
$C$: Hello.\\
$R$: Hello.\\
$U$: Hi, everyone.\\
$C$: I am Helper $C$. Nice to meet you. \\ 
$U$: Nice to meet you.\\
$C$: $U$,  is there anything you want?\\ 
$U$: I want a pen.\\
$C$: All right. $R$, Do you know where it is?\\
$R$: Yes, I know. It is on the table\\
$C$: $R$, Would you go to the table?\\
$R$: OK, I will go.\\
$R$: I am in front of the table. $U$, which pen do you want?\\
$U$: $R$, I want the pen on the right. \\
$R$: Sorry. I cannot do it. $C$, would you help me?\\
$C$: $R$, all right.  (Show the line drawn on the object by $C$ on $U$'s PC screen.) \\
$C$: I have drawn a red line on the screen. $U$, is this right? \\
$U$: Yes. OK.\\
$C$: All right.  I will make $R$ take it.\\
$R$: Done. $U$, is this OK? (The object in $R$'s hand is displayed on $U$'s PC screen.) \\
$U$: $R$, it is OK.\\
$R$: All right. I will bring it to you. ($R$ moves and arrives at $U$.)\\
$R$: $U$, here you are.\\
$U$: $R$, $C$, thank you.\\
$C$: You are welcome.\\
$R$: You are welcome.\\

\section{Statistics of SD Method}
\label{sec:appendix_dialog_SD}

\begin{table}[h]
\tbl{Statistics of the SD Method}
{
\begin{tabular}{c|cc|cc|cc}
\hline
\hline
& \multicolumn{2}{c}{M1} &  \multicolumn{2}{c}{M2}&  \multicolumn{2}{c}{M3}\\
 \hline
 No. & Mean & SE & Mean & SE & Mean & SE\\ 
 \hline
1 & 3.72 & 0.11 & 4.03 & 0.12 & 3.86 & 0.12 \\ 
  2 & 3.97 & 0.14 & 4.42 & 0.12 & 3.97 & 0.17 \\ 
  3 & 2.89 & 0.18 & 3.83 & 0.13 & 3.86 & 0.16 \\ 
  4 & 3.42 & 0.14 & 3.25 & 0.16 & 3.56 & 0.18 \\ 
  5 & 3.22 & 0.18 & 3.72 & 0.14 & 3.25 & 0.14 \\ 
  6 & 3.17 & 0.17 & 3.97 & 0.15 & 3.83 & 0.12 \\ 
  7 & 2.83 & 0.16 & 3.42 & 0.13 & 3.47 & 0.14 \\ 
  8 & 2.78 & 0.21 & 3.31 & 0.16 & 3.36 & 0.18 \\ 
  9 & 2.56 & 0.16 & 3.19 & 0.16 & 3.03 & 0.17 \\ 
  10 & 3.50 & 0.18 & 3.92 & 0.13 & 4.08 & 0.16 \\ 
  11 & 3.97 & 0.13 & 4.22 & 0.12 & 4.19 & 0.13 \\ 
  12 & 2.31 & 0.19 & 2.78 & 0.17 & 2.83 & 0.20 \\ 
  13 & 3.50 & 0.17 & 3.83 & 0.13 & 3.61 & 0.15 \\ 
  14 & 3.86 & 0.12 & 4.11 & 0.14 & 3.92 & 0.14 \\ 
  15 & 3.83 & 0.13 & 3.81 & 0.12 & 3.69 & 0.12 \\ 
  16 & 3.19 & 0.15 & 3.36 & 0.14 & 3.36 & 0.13 \\ 
  17 & 3.86 & 0.14 & 4.31 & 0.12 & 4.28 & 0.13 \\ 
  18 & 2.97 & 0.16 & 3.33 & 0.12 & 3.25 & 0.13 \\ 
  19 & 2.11 & 0.17 & 2.61 & 0.16 & 2.42 & 0.16 \\ 
  20 & 3.25 & 0.15 & 3.56 & 0.14 & 3.33 & 0.14 \\ 
  21 & 3.72 & 0.17 & 4.06 & 0.16 & 4.00 & 0.15 \\ 
  22 & 2.36 & 0.15 & 2.53 & 0.15 & 2.31 & 0.13 \\ 
  23 & 2.64 & 0.17 & 3.03 & 0.21 & 3.00 & 0.20 \\ 
   \hline

\hline
\hline
\end{tabular}
}
\label{tbl:SD_FULL}
\end{table}

\end{document}